\documentclass[11pt]{article}
\usepackage[round,authoryear]{natbib}
\PassOptionsToPackage{no-math}{fontspec}
\usepackage{metacircle}
\usepackage{float}
\usepackage{array}
\usepackage{subcaption}
\usepackage{wrapfig}
\usepackage[section,below]{placeins}

\graphicspath{{figures/}}

\newcommand{\softmax}{\operatorname{softmax}}
\newcommand{\gap}{\operatorname{gap}}
\newcommand{\code}[1]{\texttt{\detokenize{#1}}}

\title{Forking: Sudden Overfitting Under Replay}
\author{
  Shanbin Yu$^{1,3,*}$, Shaoyang Guo$^{1,4,*}$, Haoran Zhao$^{1,4}$, Danni Yu$^{1,3}$, and Ziming Liu$^{1,2,3,\dagger}$\\
  \smallskip
  \small $^1$MetaCircle (\includegraphics{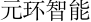}) \qquad $^2$Shanghai Qizhi Institute \qquad $^3$Tsinghua University \qquad $^4$Peking University\\
  \small $^*$Core contributors, equal contribution \quad $^\dagger$Corresponding author
}
\date{}
\paperdate{September 26, 2026}
\papertype{Meta Circle Research}
\projecturl{https://github.com/guoshaoyang-pku/forking/tree/release}

\begin{document}
\maketitle
{\renewcommand{\thefootnote}{}\footnote{Core-author contact: S. Yu, \mbox{\url{yusb24@mails.tsinghua.edu.cn}}; S. Guo, \mbox{\url{guoshaoyang@stu.pku.edu.cn}}.}\addtocounter{footnote}{-1}}

\begin{metaabstract}
This paper studies \emph{forking}, a generalization failure discovered in NanoGPT autoresearch. Under data replay, models with an over-encoding $n$-gram memory branch show a sharp separation of training and validation loss at epoch boundaries, resembling the shape of forks. We study this phenomenon in a controlled vanilla NanoGPT setting and reproduce it in a DeepSeek-style model with Engram. Mechanistically, repeated updates sharpen the continuations observed in training while suppressing the probability of unseen continuations, whose loss grows with each pass. The $n$-gram module creates weakly interacting context-specific subspaces, amplifying this effect. Low-frequency contexts contribute most of the gap, whereas larger training budgets and heavily crowded tables suppress it. We also observe forking in short-budget, heavily repeated SFT and RL-like regimes. The contributions of this paper are twofold: (1) Forking reveals yet another curious phenomenon in deep learning, in addition to grokking and double descent. (2) Forking is an unexpected and unpleasant by-product of tricks proposed by autoresearch agents. While these agents produce an enormous number of results that seem useful, we should always be careful with their results.

\end{metaabstract}

\begin{figure}[!h]
\centering
\includegraphics[width=0.94\linewidth]{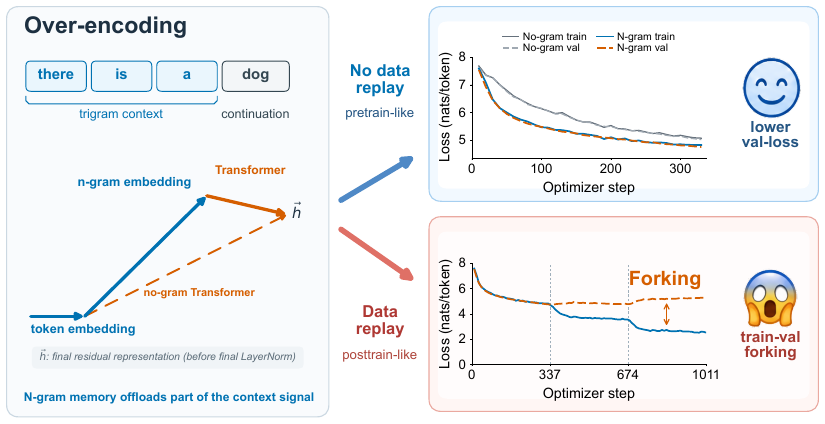}
\caption{Over-encoding and replay-aligned forking. Left: illustrative context representations. Upper right: matched models during the first pass of multi-epoch training. Lower right: the $n$-gram model under replay; vertical lines mark pass boundaries.}
\label{fig:overview}
\end{figure}
\section{Introduction}
\label{sec:introduction}

Language-model agents are increasingly used to propose, execute, and evaluate research ideas. Systems for automated scientific discovery and model improvement extend this process from individual coding tasks to iterative experimental search \citep{lu2024scientist,recursive2026research}. Such searches can improve a target metric, but they can also reveal behavior that was not part of the original objective. Forking is one such discovery from a NanoGPT autoresearch workflow (Figure~\ref{fig:overview}). The figure's schematic arrows are illustrative; the first-pass traces come from multi-epoch runs, not a separate no-replay condition.

The agent's task was to modify the model architecture and trainer to minimize next-token validation bits per byte (BPB). During a small-data experiment, an unexpected pattern occurred: training and validation losses initially improved together, but once the loader began replaying previously seen text, the curves separated rapidly near the epoch boundary. Training loss continued to fall, while validation loss stopped improving or even increased. The gap then grew in steps over subsequent passes. This replay-aligned divergence is called \emph{forking} because the two loss curves resemble a fork (Figure~\ref{fig:injection-curves}).

This behavior resembles overfitting: training loss continues to decrease while validation loss deteriorates, a pattern previously observed in language models trained on repeated data \citep{hernandez2022repeated}. However, such a sharp divergence is not an inevitable consequence of data reuse; previous studies show that language models can benefit from several passes over a finite corpus \citep{muennighoff2023scaling}. Through a series of controlled experiments, we trace forking to the interaction between data replay and \emph{over-encoding}, which enriches token representations with additional embeddings of local token combinations \citep{huang2025over}. In our model, an $n$-gram module uses the most recent $n$ tokens, including the current token, to retrieve a learned embedding from a hash-indexed table. This embedding is injected into the Transformer backbone to help predict the next token. Across the configurations we test, adding this module produces a pronounced training--validation gap once the training data are replayed.

To understand why over-encoding promotes overfitting under replay, we examine how the gap depends on context frequency. Low-frequency $n$-gram contexts account for a substantial share of the additional gap, and the average gap follows an approximate power-law relationship with context frequency over the measured range. We show that the $n$-gram table makes optimization local: by assigning distinct learned representations to different contexts, it lets the model fit each context's prediction problem independently, making it easier to memorize the continuations observed in training. For rare contexts, only a few valid continuations may be represented in the corpus. Replaying the same examples reinforces these continuations without expanding their coverage, potentially suppressing alternatives needed at validation time. We develop this explanation through frequency-dependent interventions, a model of continuation coverage, and an analysis of replay dynamics.

\noindent\textbf{Our main findings are as follows:}
\begin{enumerate}
\setlength{\itemsep}{2pt}
\setlength{\parskip}{0pt}
\item Over-encoding amplifies the divergence between training and validation loss when training data are replayed.
\item Incomplete continuation coverage and replay-driven sharpening explain why rare contexts exhibit larger gaps.
\item The gap can grow again during continued training on small replayed corpora and in a static RLVR-like proxy, even when the tables are frozen.
\end{enumerate}

The paper is organized to follow these three findings. Section~\ref{sec:phenomenon} establishes the first by recounting the discovery of forking and reproducing it in a controlled setting; Appendix~\ref{app:deepseek} extends the test to a DeepSeek-style architecture. Sections~\ref{sec:empirical} and~\ref{sec:mechanism} develop the second: the former measures the frequency dependence and tests interventions, while the latter connects these observations to continuation coverage and replay dynamics. Section~\ref{sec:transfer} examines the third through continued-training experiments and a static RLVR-like proxy, and Section~\ref{sec:discussion} discusses future directions. Due to space constraints, related work is discussed in Appendix~\ref{app:related}.

\newpage
\section{Discovery and Phenomenon}
\label{sec:phenomenon}

\noindent\begin{minipage}{\textwidth}
\subsection{Discovering and localizing forking}
\label{sec:forking-observation}
\label{sec:ophis-localization}

\setlength{\intextsep}{2pt}
\begin{wrapfigure}{r}{0.48\textwidth}
\centering
\includegraphics[width=\linewidth]{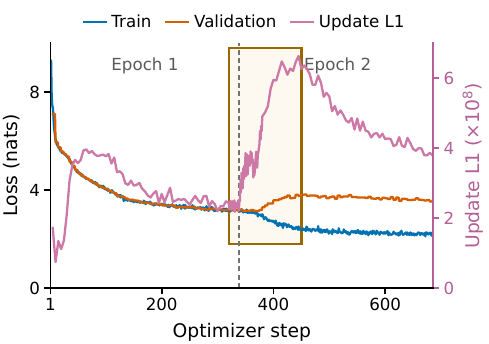}
\caption{Forking and the $n$-gram RMSProp update signal near the first replay boundary (boxed).}
\label{fig:ophis-tr1}
\end{wrapfigure}

As described in Section~\ref{sec:introduction}, forking was discovered during automated NanoGPT research: training and validation losses separate rapidly when data replay begins, and their gap grows in steps over subsequent passes. To investigate its origin, OPHIS, a method that studies training dynamics through observables of parameters, gradients, and optimizer updates \citep{yu2026ophis}, was used. Guided by OPHIS, observables were compared across replay conditions, and some $n$-gram update signals associated with the loss divergence were identified (Figure~\ref{fig:ophis-tr1}~for example). An ablation experiment was then conducted in which the bigram and trigram table contributions were disabled. The table disabling removed the rapid replay-aligned divergence, and identified the component to study. Appendix~\ref{app:screening} gives the discovery setting, screening procedure, and removal experiment.
\par\end{minipage}

\subsection{Minimal setting}
\label{sec:setting}

After the discovery-stage signal was localized to the $n$-gram pathway, we reproduced forking with NanoGPT and trainable $n$-gram tables. We take this controlled reproduction as our minimal setting. The discovery model combined several architectural and optimization choices, including gated value-path injection and multiple $n$-gram tables; the model below is simpler and distinct from the original autoresearch configuration in Appendix~\ref{app:original-setting}.

\noindent\begin{minipage}{\textwidth}
\setlength{\intextsep}{0pt}
\begin{wrapfigure}{r}{0.52\textwidth}
\centering
\includegraphics[width=\linewidth]{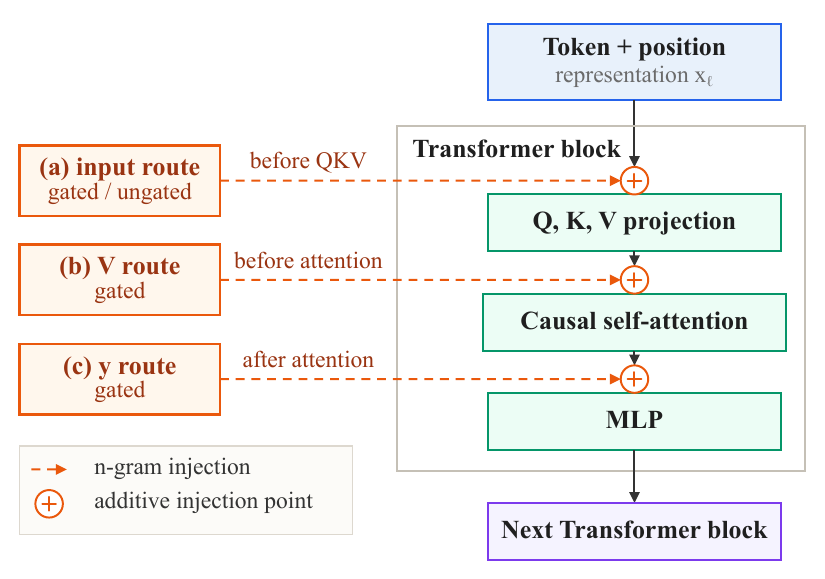}
\caption{Three $n$-gram injection routes.}
\label{fig:injection-routes}
\end{wrapfigure}

We use a NanoGPT backbone with 8 layers, 6 attention heads, a hidden dimension of 768, a vocabulary of 8192 tokens, and a sequence length of 2048, together with separate bigram and trigram lookup tables. Retrieved embeddings are injected at one of three routes---token representations before attention (input), attention values ($v$), or attention output ($y$)---alongside a no-gram baseline (Figure~\ref{fig:injection-routes}); the input route is the default. The backbone uses AdamW with learning rate $6\times10^{-4}$; the tables use momentum-free RMSProp at $128$ times the backbone learning rate. Learning rates are warmed up for 100 steps and then held constant. One epoch over the fixed single-shard training corpus takes 337 steps, and runs use seed 42 unless stated otherwise. Appendix~\ref{app:setting} gives the full contract, numerical precision, and implementation details.
\par\end{minipage}

Training loss ($L_{\mathrm{tr}}$) is the mean next-token cross-entropy on the current training batch, computed before the parameter update. Validation loss ($L_{\mathrm{va}}$) is the mean next-token cross-entropy on fixed held-out validation batches, evaluated after the update. At each evaluation step, the online gap is $G=L_{\mathrm{va}}-L_{\mathrm{tr}}$.

\begin{figure}[!htbp]
\centering
\includegraphics[width=\linewidth]{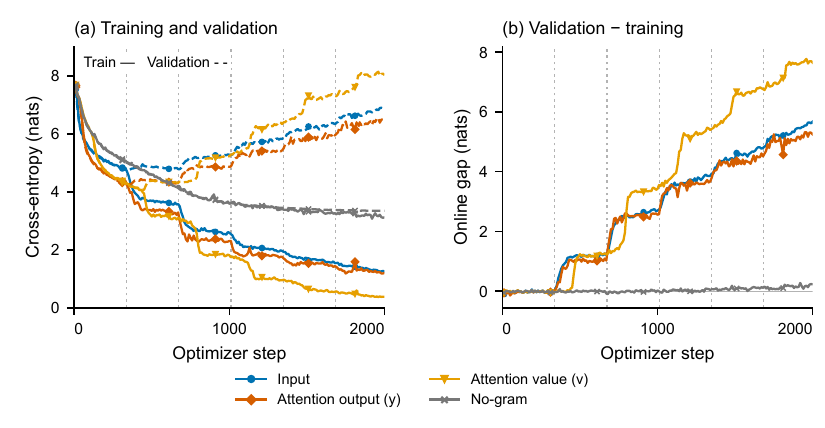}
\captionsetup{justification=centering}
\caption{Forking under data replay.}
\label{fig:injection-curves}
\end{figure}

Figure~\ref{fig:injection-curves} shows the signature pattern over 2000 updates, with dotted lines marking the 337-step replay boundaries. During the first epoch, training and validation losses improve together. Once the same text is replayed, training loss keeps falling without a corresponding validation improvement, and the gap widens---not smoothly but in a staircase aligned with the replay boundaries: most of the damage arrives right after each epoch transition and accumulates over successive passes. All three over-encoding routes fork; the attention routes ($v$, $y$) fork far more strongly than the input route, while the no-gram baseline stays small. The phenomenon is not specific to this minimal architecture: a 12-layer DeepSeek-style model with multi-layer Engram input over-encoding reproduces forking and responds to low-frequency read masking in the same way (Appendix~\ref{app:deepseek}).

\section{Empirical Laws and Interventions}
\label{sec:empirical}

\subsection{Empirical scaling laws}
\label{sec:forking-factors}

All experiments below use input injection and, where tables are enabled, a table learning rate $128$ times the backbone learning rate. The frequency analysis uses seeds 42, 43 and 44; the remaining experiments use seed 42. All use the online gap of Section~\ref{sec:setting}.

\begin{figure}[!htbp]
\centering
\begin{subfigure}{0.32\linewidth}
\centering
\includegraphics[width=\linewidth]{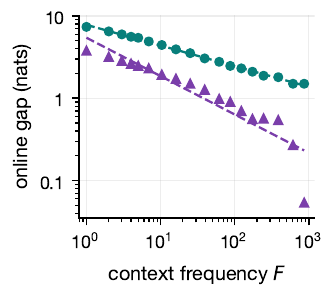}
\subcaption{}
\label{fig:frequency}
\end{subfigure}\hfill
\begin{subfigure}{0.32\linewidth}
\centering
\includegraphics[width=\linewidth]{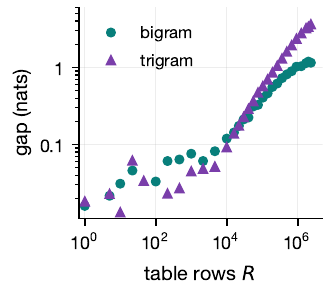}
\subcaption{}
\label{fig:capacity}
\end{subfigure}\hfill
\begin{subfigure}{0.32\linewidth}
\centering
\includegraphics[width=\linewidth]{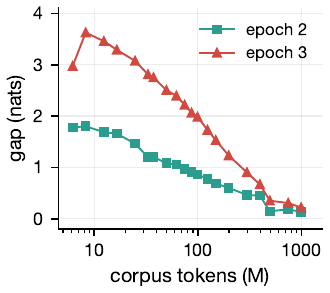}
\subcaption{}
\label{fig:corpus-size}
\end{subfigure}
\caption{Gap scaling with (a) context frequency, (b) table capacity, and (c) corpus size (trigram only). Panel (a) shows the mean online gap across seeds 42, 43 and 44 at step 1000, on log--log axes. Markers represent 18 frequency bins with $1\le F\le1000$; dashed lines are equal-bin log--log fits, with exponents $0.25$ (bigram) and $0.47$ (trigram). Appendix~\ref{app:coverage-tests} compares these with measured novelty. Panels (b,c) use seed 42 and the settings described in the text.}
\label{fig:scaling}
\end{figure}

\paragraph{Context frequency.}
Enabling both tables ($2^{20}$ rows each) and grouping the step-1000 gap by context count $F$ in the unrepeated corpus reveals a striking regularity: the bigram gap follows an approximate power law over three decades of frequency (Figure~\ref{fig:frequency}). The trigram gap also falls with frequency, with larger departures from a power law. Section~\ref{sec:coverage} connects this pattern to the fraction of novel continuations.

\paragraph{Table capacity.}
Bigger tables make forking worse. Across three decades of table rows $R$ in bigram-only and trigram-only models, the step-1000 gap generally increases, sublinearly on log-log axes, with the bigram curve flattening toward the upper end of the measured range (Figure~\ref{fig:capacity}).

\paragraph{Training corpus size.}
Bigger corpora make forking milder. In a trigram-only model with $2^{20}$ rows trained for three epochs, varying the corpus over 20 sizes from 6M to 994M prediction tokens per epoch ($0.125D_1$--$20D_1$, where $D_1$ is the single-shard budget of 337 batches), the epoch-2 and epoch-3 end gaps fall by more than an order of magnitude as the corpus grows, reaching 0.13 and 0.22 nats at $20D_1$ (Figure~\ref{fig:corpus-size}). Two factors can contribute to the decay: a context's unseen-continuation mass decreases with its training count (Appendix~\ref{app:gt-identity}), and, at fixed table size, a larger corpus spreads each table row over more distinct contexts, diluting per-context memorization. Both factors point the same way for production pretraining---a single pass over a corpus far larger than the memory---where the coverage account predicts weaker forking; the adverse corner, small corpora replayed for many passes, is exactly post-training, which we test in Section~\ref{sec:transfer}.

\begin{figure}[!ht]
\centering
\includegraphics[width=0.8\linewidth]{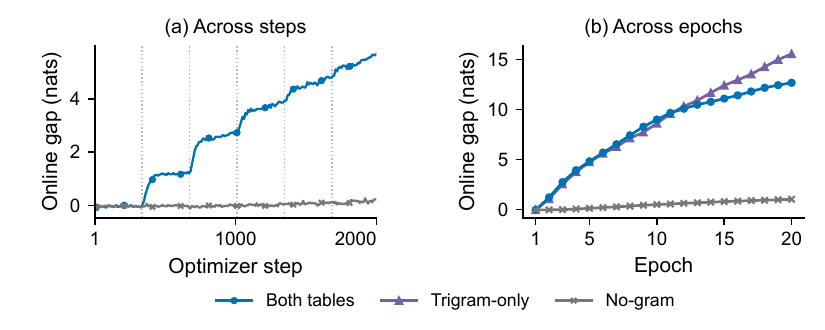}
\caption{Gap growth under replay: (a) within epochs, (b) across 20 epochs.}
\label{fig:staircase}
\end{figure}

\paragraph{Epoch count.}
With the corpus fixed, replay alone drives the growth: over 20 epochs, epoch-end gaps increase monotonically for the trigram-only and both-table models while the no-gram control's gap remains very small by comparison (Figure~\ref{fig:staircase}b). Dense 2000-step traces (Figure~\ref{fig:staircase}a) show the growth concentrating at replay boundaries, matching Figure~\ref{fig:injection-curves}b.

\subsection{Causal interventions}
\label{sec:interventions}

\begin{figure}[!ht]
\centering
\includegraphics[width=0.8\linewidth]{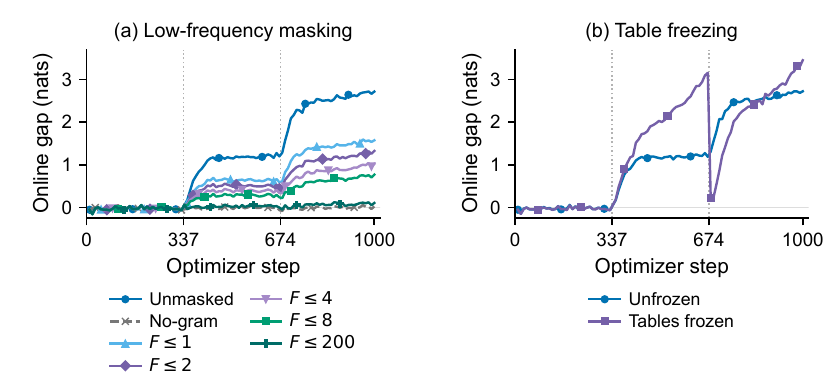}
\caption{Causal interventions after epoch 1: (a) low-frequency retrieval masking, (b) table freezing.}
\label{fig:interventions}
\end{figure}

Two interventions in the both-table model test the frequency account directly. Both use the fixed single-shard training corpus, run for 1000 updates, and start the intervention at step 338, after the first epoch; the two panels of Figure~\ref{fig:interventions} use independent runs, each with its own unmodified control.

\paragraph{Low-frequency retrieval masking.}
We count context frequencies in the unrepeated training corpus and mask the $n$-gram contribution whenever $F\leq\tau$, applying the rule separately to the bigram and trigram branches with thresholds $\tau\in\{1,2,4,8,200\}$. The retrieved embedding is set to zero during both training and evaluation, while the corresponding token remains in the training loss. Suppressing low-frequency memory access reduces the gap (Figure~\ref{fig:interventions}a): the excess gap is concentrated in rare contexts.

\paragraph{Freezing the tables.}
We train both $n$-gram tables normally during the first epoch, then freeze their parameters while continuing to update the backbone. A pronounced training--validation gap still develops after the tables are frozen (Figure~\ref{fig:interventions}b): once the tables hold learned context representations, the backbone alone can fit them during replay, so freezing the tables does not prevent forking.

\section{Mechanism of Forking}
\label{sec:mechanism}

The approximate power law in Figure~\ref{fig:frequency} is our first clue: why should context frequency organize the gap so simply?

\subsection{From the frequency law to novel continuations}
\label{sec:coverage}
A rare context has had few chances to reveal its possible continuations. Call a validation target \emph{novel} when its context occurs in training but was never followed there by that target, and let $f(F)$ be the fraction of validation positions at context count $F\ge1$ whose continuation is novel. Splitting these positions into novel and seen continuations gives the simple approximation
\begin{equation}
\gap_t(F)\approx a_t f(F)+b_t\big[1-f(F)\big],
\label{eq:coverage-main}
\end{equation}
where $a_t$ and $b_t$ are effective loss differences for novel and seen continuations, each relative to the training loss in the same frequency bin. Coverage sets the mixture weights; training determines the loss penalties. The exact decomposition allows both differences to depend on $F$ (Appendix~\ref{app:coverage}).

The novel fraction itself decays approximately as $f(F)\propto F^{-\alpha}$. For bigram contexts, fitting the same 18 frequency bins as Figure~\ref{fig:frequency} at step 1000 gives exponents $0.25$ for the three-seed mean gap and $0.23$ for measured novelty. When the novel term dominates and its penalty varies slowly, the gap inherits this decay. Good--Turing estimation and conditional continuation tails connect the novel fraction to the statistics of the text; the trigram comparison and corrections from the seen term and frequency-dependent penalties are developed in Appendix~\ref{app:coverage-tests}. The frequency law thus points to continuation coverage. It remains to explain why replay harms these continuations.

\subsection{Crowding-out under replay}
\label{sec:sharpening}
A finite corpus can underrepresent a continuation: a context whose population distribution assigns equal probability to $C$ and $D$ may appear three times with $C$ and once with $D$, so fitting the sample underpredicts $D$ on fresh data. But after the first epoch the same corpus is replayed. Its novel fraction is unchanged, while the gap continues to grow by comparable amounts over the early epochs (Figure~\ref{fig:staircase}; Appendix~\ref{app:replay-clue}). Coverage locates the vulnerable positions but does not account for this time dependence. We therefore hypothesize that each repetition reinforces seen continuations and progressively suppresses alternatives in the context-specific subspace.

This would make forking ordinary overfitting confined to very small problems. For a context $c_i$, call the next-token logits computed from it its \emph{context subspace}, and split the vocabulary into the continuations of $c_i$ seen in training, those that appear only in validation, and the rest (Appendix~\ref{app:crowding-derivation}). If training shows a single continuation $C$, the loss in that subspace is $-\ln p_C$; an idealized step directly in logit coordinates raises the logit of $C$ by $\eta(1-p_C)$ and, through the softmax, lowers every other logit by $\eta p_v$. Mass flows from the unseen continuations onto the seen one, so the seen-to-unseen odds grow by a factor above one at every epoch with nothing to stop them while $p_C<1$ (Appendix Equation~\eqref{eq:zt}). We call this \emph{crowding-out}; it provides an idealized account of the growth in the novel-continuation penalty with replay.

\begin{figure}[!ht]
\centering
\includegraphics[width=\linewidth]{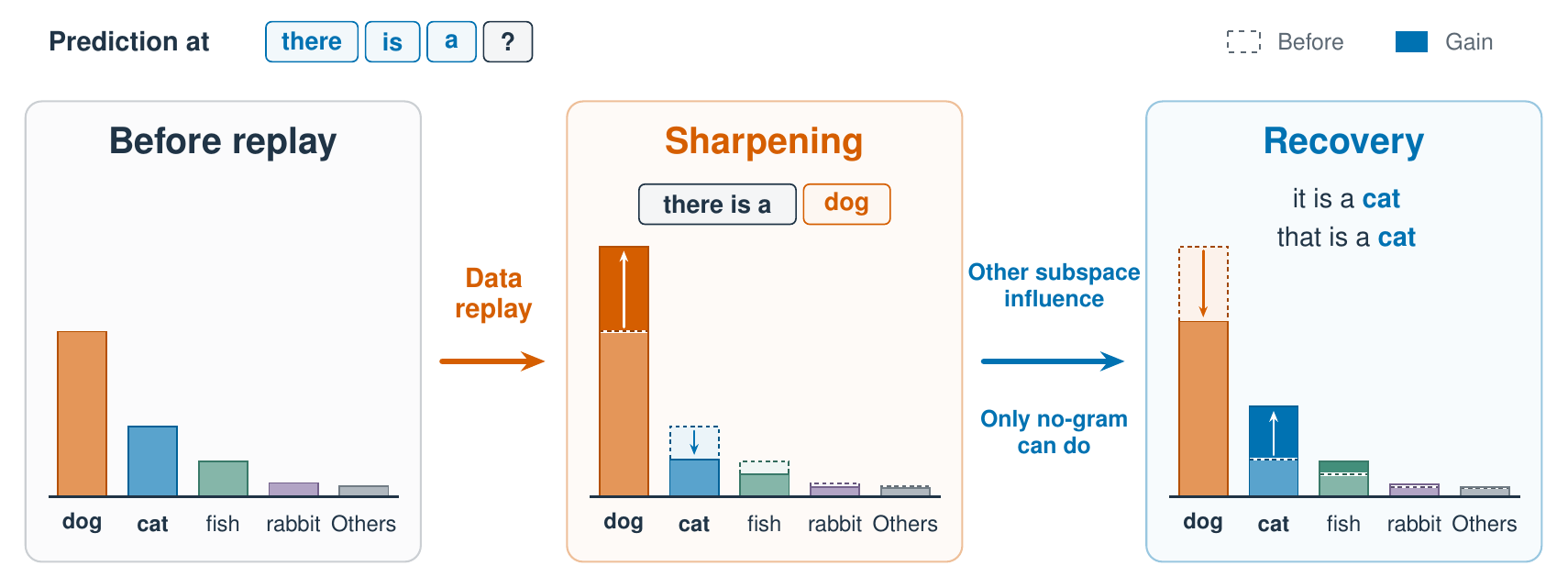}
\caption{Local sharpening and cross-context recovery, illustrated schematically. Dashed marks show prior heights; darker extensions show gains. \emph{Others} denotes the mean probability of the top 20 remaining tokens. Measured distributions for \emph{there is a} appear in Figure~\ref{fig:actual-logits-examples}(e,f).}
\label{fig:replay-transfer}
\end{figure}

Figure~\ref{fig:replay-transfer} illustrates the proposed mechanism: replay sharpens a local distribution, while updates from related contexts can support alternative continuations. Its heights are illustrative. Appendix~\ref{app:sharpening} presents the measured context distributions, including \emph{petting your}, together with aggregate seen/novel losses and predictive entropy (Figure~\ref{fig:subspace-ce}). The aggregate measurements show worsening novel-continuation accuracy in the both-table model while the no-gram control improves; individual contexts need not sharpen monotonically. Crowding-out also offers an explanation for continued forking after the tables are frozen (Section~\ref{sec:interventions}): the context-specific representation remains available, and backbone updates can continue to favor the observed continuations.

This leaves one question: why does the same replay improve the no-gram model on novel continuations, while the $n$-gram model grows less accurate? The next experiment shows that the table lets sharpening proceed independently within each context.

\FloatBarrier
\subsection{Independency of context subspaces}
\label{sec:context-locality}
Section~\ref{sec:sharpening} treats each context subspace as evolving on its own. We test this by ablation: from the start of epoch 2 or 3, we exclude the next-token loss at a random 2\% of training trigram contexts, keeping their inputs and all parameters trainable, while a disjoint 2\% serve as unmasked peers (Appendix~\ref{app:context-locality}). Once the mask starts, the masked cohort's mean gap stops growing (Figure~\ref{fig:subspace-gap}), whereas the peers' and normal training's jump again at every replay boundary. In paired one-epoch runs, masking reduces the epoch-2 gap increase of the selected contexts from $1.018$ to $0.007$ nats, while the peers' increase stays at $0.997$--$0.998$. Gap growth is generated inside each context's own subspace, not imported from the other 98\% of contexts that keep training.

\begin{figure}[!t]
\centering
\begin{subfigure}[b]{0.36\linewidth}
\centering
\includegraphics[width=\linewidth]{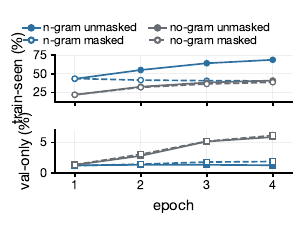}
\caption{}
\label{fig:subspace-mass}
\end{subfigure}\hfill
\begin{subfigure}[b]{0.34\linewidth}
\centering
\includegraphics[width=\linewidth]{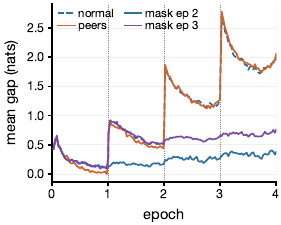}
\caption{}
\label{fig:subspace-gap}
\end{subfigure}\hfill
\begin{subfigure}[b]{0.28\linewidth}
\centering
\includegraphics[width=\linewidth]{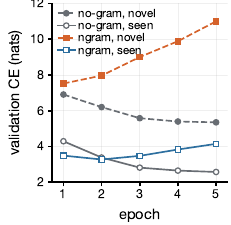}
\caption{}
\label{fig:subspace-ce}
\end{subfigure}
\caption{Context-subspace evidence for replay-driven sharpening (seed 42). (a) Mean probability mass on train-seen continuations (top) and on validation-only continuations (bottom) over four epochs, for masked and unmasked context cohorts in the $n$-gram and no-gram models. (b) Mean fixed-probe gap of the masked cohort, of its unmasked peers, and of normal training; dotted lines mark epoch boundaries. (c) Validation cross-entropy at the five epoch-end checkpoints for train-seen bigram contexts, split by whether the true continuation appeared in training.}
\label{fig:subspace-evidence}
\end{figure}

The mass on train-seen continuations shows the same partition (Figure~\ref{fig:subspace-mass}). In the $n$-gram model, unmasked contexts raise it from $0.43$ to $0.69$ by epoch 4, while masked contexts end at $0.40$. Without the $n$-gram module, masked contexts still gain mass ($0.22\to0.39$, versus $0.41$ unmasked), and replay also raises the validation-only mass ($0.013\to0.06$). The table therefore partitions optimization into weakly interacting context subspaces, inside which replay-driven sharpening proceeds unchecked by other contexts. The shared backbone still moves individual masked contexts, but it shifts their training and validation losses together and leaves their average gap flat.

\FloatBarrier

\section{Forking in Post-Training}
\label{sec:transfer}

Production pretraining rarely replays data: corpora are large enough that each example is seen once, and Section~\ref{sec:forking-factors} accordingly shows the gap shrinking as the training corpus grows. Post-training inverts both conditions. Supervised fine-tuning often replays small, expensive corpora for several epochs, and reinforcement-learning methods such as GRPO and PPO revisit the same prompts across rollout groups and optimization epochs. This section tests whether forking reappears under exactly these conditions.
\begin{figure}[!htbp]
\centering
\includegraphics[width=\linewidth]{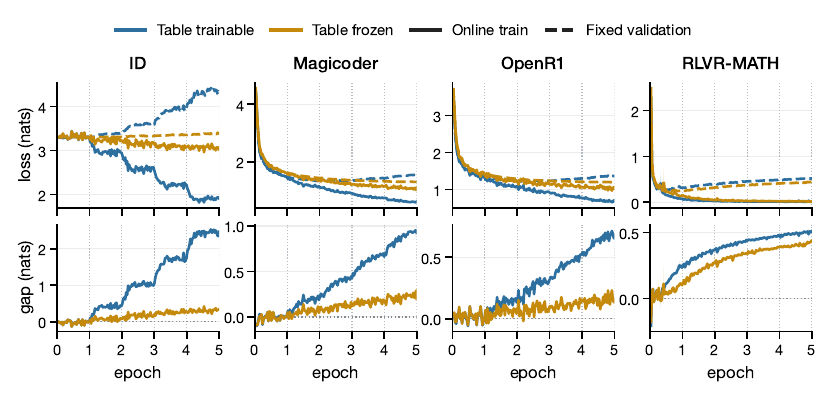}
\caption{Continued training on replayed corpora. Solid lines show online training loss, dashed lines fixed validation loss, and the lower row their difference.}
\label{fig:sft}
\end{figure}

\paragraph{Continued training on small replayed corpora.}
We start from the low-gap $20\times$-corpus trigram checkpoint (gap 0.214 at step 20{,}220), reset both optimizers with a fresh 100-step warmup, and replay three 333-batch corpora for five epochs: a same-distribution continuation of the pretraining corpus (ID), Magicoder (code) \citep{wei2024magicoder}, and OpenR1-Math (math) \citep{openr12025math}. Each corpus is run with the trigram table trainable or frozen from the start of continued training, while the backbone remains trainable. With a trainable table the gap reopens on every source: endpoint gaps are 2.463 (ID), 0.924 (Magicoder), and 0.678 (OpenR1), as shown in Figure~\ref{fig:sft}. Freezing reduces but does not eliminate the gap: the corresponding endpoints are 0.341, 0.198, and 0.158. Values for ID use seed 42; the other sources average downstream seeds 42, 43, and 44 from the same pretrained checkpoint. The full protocol and endpoint table are in Appendix~\ref{app:sft}.

\paragraph{Verifier-selected data as an RL proxy.}
Verifier-filtered policy optimization revisits prompts across rollout groups and optimization epochs. As a data-reuse proxy, we replay RLVR-MATH from the Tulu 3 release~\citep{lambert2024tulu3}, with the reference answer appended once, using an all-token language-model objective rather than on-policy reinforcement learning. The prepared 333-batch pool contains approximately 8.54 raw-source token cycles per nominal epoch. Mean endpoint gaps are 0.509 (trainable) and 0.434 (frozen), as shown in the RLVR-MATH column of Figure~\ref{fig:sft}; the frozen-table residual exceeds that on the three other sources.

These continued-training experiments show that replaying a narrow corpus can reopen a train--validation gap, and freezing the local memory mitigates but does not prevent it. They motivate monitoring this divergence in replay-heavy post-training pipelines.

\section{Future Work}
\label{sec:discussion}

Forking under over-encoding with data replay is a robust phenomenon, and its impact is largest in the short-budget, heavily repeated regimes of SFT and RL. This paper concentrates on the mechanism, which we consider reasonably complete; the next step is to characterize forking in production-scale models and to determine which production failures it accounts for. Such a study is currently limited by the scarcity of open over-encoding models---the most prominent, DeepSeek-V4.1-Flash \citep{deepseek2026v41flash}, is very recent and very large---but its importance will grow as over-encoding matures.

Beyond loss regularities, future work should locate the downstream performance harm of forking and develop ways to avoid it. Masking low-frequency contexts is a promising candidate. An alternative is to restrict the over-encoding vocabulary at construction time by BPE-style aggregation, extrapolating the tokenizer-training algorithm so that low-frequency n-grams never enter the table; the over-encoding scheme of SCONE \citep{yu2025scone} already adopts this design.

\bibliography{references}

\begin{thebibliography}{54}
\providecommand{\natexlab}[1]{#1}
\providecommand{\url}[1]{\texttt{#1}}
\expandafter\ifx\csname urlstyle\endcsname\relax
  \providecommand{\doi}[1]{doi: #1}\else
  \providecommand{\doi}{doi: \begingroup \urlstyle{rm}\Url}\fi

\bibitem[Berges et~al.(2025)Berges, Oguz, Haziza, Yih, Zettlemoyer, and
  Ghosh]{berges2025memory}
Vincent-Pierre Berges, Barlas Oguz, Daniel Haziza, Wen-Tau Yih, Luke
  Zettlemoyer, and Gargi Ghosh.
\newblock Memory layers at scale.
\newblock In \emph{Proceedings of the 42nd International Conference on Machine
  Learning}, volume 267 of \emph{Proceedings of Machine Learning Research},
  pages 3831--3842. PMLR, 2025.
\newblock URL \url{https://proceedings.mlr.press/v267/berges25a.html}.

\bibitem[Biderman et~al.(2023)Biderman, Prashanth, Sutawika, Schoelkopf,
  Anthony, Purohit, and Raff]{biderman2023emergent}
Stella Biderman, USVSN~Sai Prashanth, Lintang Sutawika, Hailey Schoelkopf,
  Quentin Anthony, Shivanshu Purohit, and Edward Raff.
\newblock Emergent and predictable memorization in large language models.
\newblock In \emph{Advances in Neural Information Processing Systems},
  volume~36, pages 28072--28090. Curran Associates, Inc., 2023.
\newblock URL
  \url{https://proceedings.neurips.cc/paper_files/paper/2023/hash/59404fb89d6194641c69ae99ecdf8f6d-Abstract-Conference.html}.

\bibitem[Bojanowski et~al.(2017)Bojanowski, Grave, Joulin, and
  Mikolov]{bojanowski2017subword}
Piotr Bojanowski, Edouard Grave, Armand Joulin, and Tomas Mikolov.
\newblock Enriching word vectors with subword information.
\newblock \emph{Transactions of the Association for Computational Linguistics},
  5:\penalty0 135--146, 2017.
\newblock \doi{10.1162/tacl_a_00051}.
\newblock URL \url{https://aclanthology.org/Q17-1010/}.

\bibitem[Borgeaud et~al.(2022)Borgeaud, Mensch, Hoffmann, Cai, Rutherford,
  Millican, van~den Driessche, Lespiau, Damoc, Clark, de~Las~Casas, Guy,
  Menick, Ring, Hennigan, Huang, Maggiore, Jones, Cassirer, Brock, Paganini,
  Irving, Vinyals, Osindero, Simonyan, Rae, Elsen, and
  Sifre]{borgeaud2021retro}
Sebastian Borgeaud, Arthur Mensch, Jordan Hoffmann, Trevor Cai, Eliza
  Rutherford, Katie Millican, George van~den Driessche, Jean-Baptiste Lespiau,
  Bogdan Damoc, Aidan Clark, Diego de~Las~Casas, Aurelia Guy, Jacob Menick,
  Roman Ring, Tom Hennigan, Saffron Huang, Loren Maggiore, Chris Jones, Albin
  Cassirer, Andy Brock, Michela Paganini, Geoffrey Irving, Oriol Vinyals, Simon
  Osindero, Karen Simonyan, Jack~W. Rae, Erich Elsen, and Laurent Sifre.
\newblock Improving language models by retrieving from trillions of tokens.
\newblock In \emph{Proceedings of the 39th International Conference on Machine
  Learning}, volume 162 of \emph{Proceedings of Machine Learning Research},
  pages 2206--2240. PMLR, 2022.
\newblock URL \url{https://proceedings.mlr.press/v162/borgeaud22a.html}.

\bibitem[Carlini et~al.(2023)Carlini, Ippolito, Jagielski, Lee, Tram{\`e}r, and
  Zhang]{carlini2023quantifying}
Nicholas Carlini, Daphne Ippolito, Matthew Jagielski, Katherine Lee, Florian
  Tram{\`e}r, and Chiyuan Zhang.
\newblock Quantifying memorization across neural language models.
\newblock In \emph{International Conference on Learning Representations}, 2023.
\newblock URL \url{https://openreview.net/forum?id=TatRHT_1cK}.

\bibitem[Carlsson et~al.(2025)Carlsson, Liu, Ward, Kurfali, and
  Nivre]{carlsson2025hyperfitting}
Fredrik Carlsson, Fangyu Liu, Daniel Ward, Murathan Kurfali, and Joakim Nivre.
\newblock The hyperfitting phenomenon: Sharpening and stabilizing {LLMs} for
  open-ended text generation.
\newblock In \emph{International Conference on Learning Representations}, 2025.
\newblock URL \url{https://openreview.net/forum?id=Ij9ilPh36h}.

\bibitem[Chen and Goodman(1999)]{chen1999smoothing}
Stanley~F. Chen and Joshua Goodman.
\newblock An empirical study of smoothing techniques for language modeling.
\newblock \emph{Computer Speech \& Language}, 13\penalty0 (4):\penalty0
  359--394, 1999.
\newblock \doi{10.1006/csla.1999.0128}.
\newblock URL \url{https://doi.org/10.1006/csla.1999.0128}.

\bibitem[Cheng et~al.(2026{\natexlab{a}})Cheng, Guan, Wei, Sun, Li, Du, Xiong,
  Yuan, Lu, and Gong]{cheng2026memorygrafting}
Runxi Cheng, Yuchen Guan, Yongxian Wei, Qianpu Sun, Qixiu Li, Sinan Du, Feng
  Xiong, Chun Yuan, Yan Lu, and Yeyun Gong.
\newblock {Memory Grafting}: Scaling language model pre-training via offline
  conditional memory, 2026{\natexlab{a}}.
\newblock URL \url{https://arxiv.org/abs/2605.20948}.

\bibitem[Cheng et~al.(2026{\natexlab{b}})Cheng, Zeng, Dai, Chen, Wang, Xie,
  Huang, Yu, Hao, Zhang, Li, Zhang, Zhao, and Liang]{cheng2026engram}
Xin Cheng, Wangding Zeng, Damai Dai, Qinyu Chen, Bingxuan Wang, Zhenda Xie,
  Kezhao Huang, Xingkai Yu, Zhewen Hao, Han Zhang, Yu-Kun Li, Huishuai Zhang,
  Dongyan Zhao, and Wenfeng Liang.
\newblock Conditional memory via scalable lookup: A new axis of sparsity for
  large language models.
\newblock In Maria Liakata, Viviane~P. Moreira, Jiajun Zhang, and David
  Jurgens, editors, \emph{Proceedings of the 64th Annual Meeting of the
  {A}ssociation for {C}omputational {L}inguistics (Volume 1: Long Papers)},
  pages 4968--4990, San Diego, California, United States, July
  2026{\natexlab{b}}. Association for Computational Linguistics.
\newblock ISBN 979-8-89176-390-6.
\newblock \doi{10.18653/v1/2026.acl-long.226}.
\newblock URL \url{https://aclanthology.org/2026.acl-long.226/}.

\bibitem[{DeepSeek-AI} et~al.(2026)]{deepseek2026v41flash}
{DeepSeek-AI} et~al.
\newblock {DeepSeek-V4.1-Flash}: Pushing the limits of {KV} cache compression,
  2026.
\newblock URL \url{https://arxiv.org/abs/2609.19969}.

\bibitem[Feldman(2020)]{feldman2020memorization}
Vitaly Feldman.
\newblock Does learning require memorization? {A} short tale about a long tail.
\newblock In \emph{Proceedings of the 52nd Annual {ACM} {SIGACT} Symposium on
  Theory of Computing ({STOC})}, pages 954--959. ACM, 2020.
\newblock \doi{10.1145/3357713.3384290}.
\newblock URL \url{https://doi.org/10.1145/3357713.3384290}.

\bibitem[Feldman and Zhang(2020)]{feldman2020longtail}
Vitaly Feldman and Chiyuan Zhang.
\newblock What neural networks memorize and why: Discovering the long tail via
  influence estimation.
\newblock In \emph{Advances in Neural Information Processing Systems},
  volume~33, pages 2881--2891. Curran Associates, Inc., 2020.
\newblock URL
  \url{https://proceedings.neurips.cc/paper_files/paper/2020/hash/1e14bfe2714193e7af5abc64ecbd6b46-Abstract.html}.

\bibitem[Gale and Sampson(1995)]{gale1995good}
William~A. Gale and Geoffrey Sampson.
\newblock {Good--Turing} frequency estimation without tears.
\newblock \emph{Journal of Quantitative Linguistics}, 2\penalty0 (3):\penalty0
  217--237, 1995.
\newblock \doi{10.1080/09296179508590051}.
\newblock URL \url{https://www.grsampson.net/AGTf.pdf}.

\bibitem[Geva et~al.(2021)Geva, Schuster, Berant, and Levy]{geva2021ffn}
Mor Geva, Roei Schuster, Jonathan Berant, and Omer Levy.
\newblock {Transformer} feed-forward layers are key-value memories.
\newblock In \emph{Proceedings of the 2021 Conference on Empirical Methods in
  Natural Language Processing}, pages 5484--5495, 2021.
\newblock \doi{10.18653/v1/2021.emnlp-main.446}.
\newblock URL \url{https://doi.org/10.18653/v1/2021.emnlp-main.446}.

\bibitem[Gnedin et~al.(2007)Gnedin, Hansen, and Pitman]{gnedin2007occupancy}
Alexander Gnedin, Ben Hansen, and Jim Pitman.
\newblock Notes on the occupancy problem with infinitely many boxes: General
  asymptotics and power laws.
\newblock \emph{Probability Surveys}, 4:\penalty0 146--171, 2007.
\newblock \doi{10.1214/07-PS092}.
\newblock URL \url{https://arxiv.org/abs/math/0701718}.

\bibitem[Good(1953)]{good1953population}
I.~J. Good.
\newblock The population frequencies of species and the estimation of
  population parameters.
\newblock \emph{Biometrika}, 40\penalty0 (3--4):\penalty0 237--264, 1953.
\newblock \doi{10.1093/biomet/40.3-4.237}.

\bibitem[Grave et~al.(2017)Grave, Joulin, and Usunier]{grave2016cache}
Edouard Grave, Armand Joulin, and Nicolas Usunier.
\newblock Improving neural language models with a continuous cache.
\newblock In \emph{International Conference on Learning Representations}, 2017.
\newblock URL \url{https://openreview.net/forum?id=B184E5qee}.

\bibitem[Guo et~al.(2017)Guo, Pleiss, Sun, and Weinberger]{guo2017calibration}
Chuan Guo, Geoff Pleiss, Yu~Sun, and Kilian~Q. Weinberger.
\newblock On calibration of modern neural networks.
\newblock In \emph{Proceedings of the 34th International Conference on Machine
  Learning}, volume~70 of \emph{Proceedings of Machine Learning Research},
  pages 1321--1330. PMLR, 2017.
\newblock URL \url{https://proceedings.mlr.press/v70/guo17a.html}.

\bibitem[Hernandez et~al.(2022)Hernandez, Brown, Conerly, DasSarma, Drain,
  El-Showk, Elhage, Hatfield-Dodds, Henighan, Hume, Johnston, Mann, Olah,
  Olsson, Amodei, Joseph, Kaplan, and McCandlish]{hernandez2022repeated}
Danny Hernandez, Tom Brown, Tom Conerly, Nova DasSarma, Dawn Drain, Sheer
  El-Showk, Nelson Elhage, Zac Hatfield-Dodds, Tom Henighan, Tristan Hume,
  Scott Johnston, Ben Mann, Chris Olah, Catherine Olsson, Dario Amodei,
  Nicholas Joseph, Jared Kaplan, and Sam McCandlish.
\newblock Scaling laws and interpretability of learning from repeated data.
\newblock \emph{arXiv preprint arXiv:2205.10487}, 2022.
\newblock URL \url{https://arxiv.org/abs/2205.10487}.

\bibitem[Howard and Whitaker(2023)]{howard2023singleexample}
Jeremy Howard and Jonathan Whitaker.
\newblock Can {LLMs} learn from a single example?
\newblock fast.ai blog, September 2023.
\newblock URL \url{https://www.fast.ai/posts/2023-09-04-learning-jumps/}.
\newblock Published September 4, 2023.

\bibitem[Huang et~al.(2025)Huang, Zhu, Wu, Zeng, Wang, Min, and
  Xun]{huang2025over}
Hongzhi Huang, Defa Zhu, Banggu Wu, Yutao Zeng, Ya~Wang, Qiyang Min, and Zhou
  Xun.
\newblock {Over-Tokenized Transformer}: Vocabulary is generally worth scaling.
\newblock In \emph{Proceedings of the 42nd International Conference on Machine
  Learning}, volume 267 of \emph{Proceedings of Machine Learning Research},
  pages 26261--26282. PMLR, 13--19 Jul 2025.
\newblock URL \url{https://proceedings.mlr.press/v267/huang25bb.html}.

\bibitem[Huang et~al.(2024)Huang, Vora, Liang, and
  Leskovec]{huang2024mlagentbench}
Qian Huang, Jian Vora, Percy Liang, and Jure Leskovec.
\newblock {MLAgentBench}: Evaluating language agents on machine learning
  experimentation.
\newblock In \emph{Proceedings of the 41st International Conference on Machine
  Learning}, volume 235 of \emph{Proceedings of Machine Learning Research},
  pages 20271--20309. PMLR, 2024.
\newblock URL \url{https://proceedings.mlr.press/v235/huang24y.html}.

\bibitem[{Hugging Face}(2025)]{openr12025math}
{Hugging Face}.
\newblock {OpenR1-Math-220k}.
\newblock Hugging Face dataset, 2025.
\newblock URL \url{https://huggingface.co/datasets/open-r1/OpenR1-Math-220k}.

\bibitem[Jha et~al.(2026)Jha, Li, Leskovec, Liang, and
  Zettlemoyer]{jha2026datascarcity}
Atindra Jha, Margaret Li, Jure Leskovec, Percy Liang, and Luke Zettlemoyer.
\newblock Data scarcity and model sparsity: Mixtures-of-experts overfit more to
  repeated data, 2026.
\newblock URL \url{https://arxiv.org/abs/2609.11917}.

\bibitem[Kandpal et~al.(2022)Kandpal, Wallace, and Raffel]{kandpal2022dedup}
Nikhil Kandpal, Eric Wallace, and Colin Raffel.
\newblock Deduplicating training data mitigates privacy risks in language
  models.
\newblock In \emph{Proceedings of the 39th International Conference on Machine
  Learning}, volume 162 of \emph{Proceedings of Machine Learning Research},
  pages 10697--10707. PMLR, 2022.
\newblock URL \url{https://proceedings.mlr.press/v162/kandpal22a.html}.

\bibitem[Kandpal et~al.(2023)Kandpal, Deng, Roberts, Wallace, and
  Raffel]{kandpal2023longtail}
Nikhil Kandpal, Haikang Deng, Adam Roberts, Eric Wallace, and Colin Raffel.
\newblock Large language models struggle to learn long-tail knowledge.
\newblock In \emph{Proceedings of the 40th International Conference on Machine
  Learning}, volume 202 of \emph{Proceedings of Machine Learning Research},
  pages 15696--15707. PMLR, 2023.
\newblock URL \url{https://proceedings.mlr.press/v202/kandpal23a.html}.

\bibitem[Karpathy(2022)]{karpathy2022nanogpt}
Andrej Karpathy.
\newblock {nanoGPT}.
\newblock GitHub repository, 2022.
\newblock URL \url{https://github.com/karpathy/nanoGPT}.
\newblock Accessed 2026-09-16.

\bibitem[Karpathy(2026)]{karpathy2026autoresearch}
Andrej Karpathy.
\newblock Autoresearch.
\newblock GitHub repository, 2026.
\newblock URL \url{https://github.com/karpathy/autoresearch}.
\newblock Accessed 2026-09-16.

\bibitem[Katz(1987)]{katz1987estimation}
Slava~M. Katz.
\newblock Estimation of probabilities from sparse data for the language model
  component of a speech recognizer.
\newblock \emph{IEEE Transactions on Acoustics, Speech, and Signal Processing},
  35\penalty0 (3):\penalty0 400--401, 1987.
\newblock \doi{10.1109/TASSP.1987.1165125}.
\newblock URL \url{https://doi.org/10.1109/TASSP.1987.1165125}.

\bibitem[Khandelwal et~al.(2020)Khandelwal, Levy, Jurafsky, Zettlemoyer, and
  Lewis]{khandelwal2019knnlm}
Urvashi Khandelwal, Omer Levy, Dan Jurafsky, Luke Zettlemoyer, and Mike Lewis.
\newblock Generalization through memorization: Nearest neighbor language
  models.
\newblock In \emph{International Conference on Learning Representations}, 2020.
\newblock URL \url{https://openreview.net/forum?id=HklBjCEKvH}.

\bibitem[Kneser and Ney(1995)]{kneser1995backing}
Reinhard Kneser and Hermann Ney.
\newblock Improved backing-off for {M-gram} language modeling.
\newblock In \emph{Proceedings of the 1995 International Conference on
  Acoustics, Speech, and Signal Processing ({ICASSP})}, volume~1, pages
  181--184. IEEE, 1995.
\newblock \doi{10.1109/ICASSP.1995.479394}.
\newblock URL \url{https://doi.org/10.1109/ICASSP.1995.479394}.

\bibitem[Lambert et~al.(2025)Lambert, Morrison, Pyatkin, Huang, Ivison,
  Brahman, Miranda, Liu, Dziri, Lyu, Gu, Malik, Graf, Hwang, Yang, Bras,
  Tafjord, Wilhelm, Soldaini, Smith, Wang, Dasigi, and
  Hajishirzi]{lambert2024tulu3}
Nathan Lambert, Jacob Morrison, Valentina Pyatkin, Shengyi Huang, Hamish
  Ivison, Faeze Brahman, Lester James~Validad Miranda, Alisa Liu, Nouha Dziri,
  Xinxi Lyu, Yuling Gu, Saumya Malik, Victoria Graf, Jena~D. Hwang, Jiangjiang
  Yang, Ronan~Le Bras, Oyvind Tafjord, Christopher Wilhelm, Luca Soldaini,
  Noah~A. Smith, Yizhong Wang, Pradeep Dasigi, and Hannaneh Hajishirzi.
\newblock {T{\"U}LU 3}: Pushing frontiers in open language model post-training.
\newblock In \emph{Second Conference on Language Modeling}, 2025.
\newblock URL \url{https://openreview.net/forum?id=i1uGbfHHpH}.

\bibitem[Lample et~al.(2019)Lample, Sablayrolles, Ranzato, Denoyer, and
  J{\'e}gou]{lample2019product}
Guillaume Lample, Alexandre Sablayrolles, Marc'Aurelio Ranzato, Ludovic
  Denoyer, and Herv{\'e} J{\'e}gou.
\newblock Large memory layers with product keys.
\newblock In \emph{Advances in Neural Information Processing Systems},
  volume~32. Curran Associates, Inc., 2019.
\newblock URL
  \url{https://proceedings.neurips.cc/paper_files/paper/2019/file/9d8df73a3cfbf3c5b47bc9b50f214aff-Paper.pdf}.

\bibitem[Lee et~al.(2022)Lee, Ippolito, Nystrom, Zhang, Eck, Callison-Burch,
  and Carlini]{lee2022dedup}
Katherine Lee, Daphne Ippolito, Andrew Nystrom, Chiyuan Zhang, Douglas Eck,
  Chris Callison-Burch, and Nicholas Carlini.
\newblock Deduplicating training data makes language models better.
\newblock In \emph{Proceedings of the 60th Annual Meeting of the Association
  for Computational Linguistics (Volume 1: Long Papers)}, pages 8424--8445,
  2022.
\newblock \doi{10.18653/v1/2022.acl-long.577}.
\newblock URL \url{https://doi.org/10.18653/v1/2022.acl-long.577}.

\bibitem[Liu et~al.(2024)Liu, Min, Zettlemoyer, Choi, and
  Hajishirzi]{liu2024infinigram}
Jiacheng Liu, Sewon Min, Luke Zettlemoyer, Yejin Choi, and Hannaneh Hajishirzi.
\newblock {Infini-gram}: Scaling unbounded n-gram language models to a trillion
  tokens.
\newblock In \emph{First Conference on Language Modeling ({COLM})}, 2024.
\newblock URL \url{https://openreview.net/forum?id=u2vAyMeLMm}.

\bibitem[Lu et~al.(2024)Lu, Lu, Lange, Foerster, Clune, and
  Ha]{lu2024scientist}
Chris Lu, Cong Lu, Robert~Tjarko Lange, Jakob Foerster, Jeff Clune, and David
  Ha.
\newblock The {AI Scientist}: Towards fully automated open-ended scientific
  discovery.
\newblock \emph{arXiv preprint arXiv:2408.06292}, 2024.
\newblock URL \url{https://arxiv.org/abs/2408.06292}.

\bibitem[Lu et~al.(2026)Lu, Lu, Lange, Yamada, Hu, Foerster, Ha, and
  Clune]{lu2026nature}
Chris Lu, Cong Lu, Robert~Tjarko Lange, Yutaro Yamada, Shengran Hu, Jakob
  Foerster, David Ha, and Jeff Clune.
\newblock Towards end-to-end automation of {AI} research.
\newblock \emph{Nature}, 651\penalty0 (8107):\penalty0 914--919, 2026.
\newblock \doi{10.1038/s41586-026-10265-5}.
\newblock URL \url{https://doi.org/10.1038/s41586-026-10265-5}.

\bibitem[Muennighoff et~al.(2023)Muennighoff, Rush, Barak, {Le Scao}, Piktus,
  Tazi, Pyysalo, Wolf, and Raffel]{muennighoff2023scaling}
Niklas Muennighoff, Alexander~M. Rush, Boaz Barak, Teven {Le Scao}, Aleksandra
  Piktus, Nouamane Tazi, Sampo Pyysalo, Thomas Wolf, and Colin Raffel.
\newblock Scaling data-constrained language models.
\newblock In \emph{Advances in Neural Information Processing Systems},
  volume~36, pages 50358--50376. Curran Associates, Inc., 2023.
\newblock URL
  \url{https://proceedings.neurips.cc/paper_files/paper/2023/file/9d89448b63ce1e2e8dc7af72c984c196-Paper-Conference.pdf}.

\bibitem[Pagnoni et~al.(2025)Pagnoni, Pasunuru, Rodriguez, Nguyen, Muller, Li,
  Zhou, Yu, Weston, Zettlemoyer, Ghosh, Lewis, Holtzman, and
  Iyer]{pagnoni2025blt}
Artidoro Pagnoni, Ramakanth Pasunuru, Pedro Rodriguez, John Nguyen, Benjamin
  Muller, Margaret Li, Chunting Zhou, Lili Yu, Jason~E. Weston, Luke
  Zettlemoyer, Gargi Ghosh, Mike Lewis, Ari Holtzman, and Srini Iyer.
\newblock {Byte Latent Transformer}: Patches scale better than tokens.
\newblock In \emph{Proceedings of the 63rd Annual Meeting of the Association
  for Computational Linguistics (Volume 1: Long Papers)}, pages 9238--9258,
  2025.
\newblock \doi{10.18653/v1/2025.acl-long.453}.
\newblock URL \url{https://doi.org/10.18653/v1/2025.acl-long.453}.

\bibitem[{Recursive}(2026)]{recursive2026research}
{Recursive}.
\newblock First steps toward automated {AI} research.
\newblock Blog post, June 2026.
\newblock URL
  \url{https://www.recursive.com/articles/first-steps-toward-automated-ai-research}.

\bibitem[Roy et~al.(2022)Roy, Anil, Lai, Lee, Zhao, Zhang, Wang, Zhang, Wu,
  Swavely, Yu, Dao, Fifty, Chen, and Wu]{roy2022ngrammer}
Aurko Roy, Rohan Anil, Guangda Lai, Benjamin Lee, Jeffrey Zhao, Shuyuan Zhang,
  Shibo Wang, Ye~Zhang, Shen Wu, Rigel Swavely, Tao~(Alex) Yu, Phuong Dao,
  Christopher Fifty, Zhifeng Chen, and Yonghui Wu.
\newblock {N-Grammer}: Augmenting {Transformers} with latent n-grams, 2022.
\newblock URL \url{https://arxiv.org/abs/2207.06366}.

\bibitem[Svenstrup et~al.(2017)Svenstrup, Hansen, and
  Winther]{svenstrup2017hash}
Dan Svenstrup, Jonas~Meinertz Hansen, and Ole Winther.
\newblock Hash embeddings for efficient word representations.
\newblock In \emph{Advances in Neural Information Processing Systems},
  volume~30. Curran Associates, Inc., 2017.
\newblock URL
  \url{https://proceedings.neurips.cc/paper_files/paper/2017/file/f0f6ba4b5e0000340312d33c212c3ae8-Paper.pdf}.

\bibitem[Teh(2006)]{teh2006hierarchical}
Yee~Whye Teh.
\newblock A hierarchical {B}ayesian language model based on {P}itman-{Y}or
  processes.
\newblock In Nicoletta Calzolari, Claire Cardie, and Pierre Isabelle, editors,
  \emph{Proceedings of the 21st International Conference on Computational
  Linguistics and 44th Annual Meeting of the Association for Computational
  Linguistics}, pages 985--992, Sydney, Australia, July 2006. Association for
  Computational Linguistics.
\newblock \doi{10.3115/1220175.1220299}.
\newblock URL \url{https://aclanthology.org/P06-1124/}.

\bibitem[Tirumala et~al.(2022)Tirumala, Markosyan, Zettlemoyer, and
  Aghajanyan]{tirumala2022memorization}
Kushal Tirumala, Aram~H. Markosyan, Luke Zettlemoyer, and Armen Aghajanyan.
\newblock Memorization without overfitting: Analyzing the training dynamics of
  large language models.
\newblock In \emph{Advances in Neural Information Processing Systems},
  volume~35, pages 38274--38290. Curran Associates, Inc., 2022.
\newblock URL
  \url{https://proceedings.neurips.cc/paper_files/paper/2022/file/fa0509f4dab6807e2cb465715bf2d249-Paper-Conference.pdf}.

\bibitem[Wei et~al.(2024)Wei, Wang, Liu, Ding, and Zhang]{wei2024magicoder}
Yuxiang Wei, Zhe Wang, Jiawei Liu, Yifeng Ding, and Lingming Zhang.
\newblock {Magicoder}: Empowering code generation with {OSS-Instruct}.
\newblock In \emph{Proceedings of the 41st International Conference on Machine
  Learning}, volume 235 of \emph{Proceedings of Machine Learning Research},
  pages 52632--52657. PMLR, 2024.
\newblock URL \url{https://proceedings.mlr.press/v235/wei24h.html}.

\bibitem[Weinberger et~al.(2009)Weinberger, Dasgupta, Langford, Smola, and
  Attenberg]{weinberger2009hashing}
Kilian~Q. Weinberger, Anirban Dasgupta, John Langford, Alexander~J. Smola, and
  Josh Attenberg.
\newblock Feature hashing for large scale multitask learning.
\newblock In \emph{Proceedings of the 26th Annual International Conference on
  Machine Learning ({ICML})}, pages 1113--1120. ACM, 2009.
\newblock \doi{10.1145/1553374.1553516}.
\newblock URL \url{https://doi.org/10.1145/1553374.1553516}.

\bibitem[Wu et~al.(2022)Wu, Rabe, Hutchins, and Szegedy]{wu2022memorizing}
Yuhuai Wu, Markus~N. Rabe, DeLesley Hutchins, and Christian Szegedy.
\newblock {Memorizing Transformers}.
\newblock In \emph{International Conference on Learning Representations}, 2022.
\newblock URL \url{https://openreview.net/forum?id=TrjbxzRcnf-}.

\bibitem[Xu et~al.(2023)Xu, Alon, and Neubig]{xu2023whyknn}
Frank~F. Xu, Uri Alon, and Graham Neubig.
\newblock Why do nearest neighbor language models work?
\newblock In \emph{Proceedings of the 40th International Conference on Machine
  Learning}, volume 202 of \emph{Proceedings of Machine Learning Research},
  pages 38325--38341. PMLR, 2023.
\newblock URL \url{https://proceedings.mlr.press/v202/xu23a.html}.

\bibitem[Xue et~al.(2023)Xue, Fu, Zhou, Zheng, and You]{xue2023repeat}
Fuzhao Xue, Yao Fu, Wangchunshu Zhou, Zangwei Zheng, and Yang You.
\newblock To repeat or not to repeat: Insights from scaling {LLM} under
  token-crisis.
\newblock In \emph{Advances in Neural Information Processing Systems},
  volume~36, pages 59304--59322. Curran Associates, Inc., 2023.
\newblock URL
  \url{https://proceedings.neurips.cc/paper_files/paper/2023/hash/b9e472cd579c83e2f6aa3459f46aac28-Abstract-Conference.html}.

\bibitem[Yamada et~al.(2025)Yamada, Lange, Lu, Hu, Lu, Foerster, Clune, and
  Ha]{yamada2025scientistv2}
Yutaro Yamada, Robert~Tjarko Lange, Cong Lu, Shengran Hu, Chris Lu, Jakob
  Foerster, Jeff Clune, and David Ha.
\newblock The {AI Scientist-v2}: Workshop-level automated scientific discovery
  via agentic tree search.
\newblock arXiv preprint arXiv:2504.08066, 2025.
\newblock URL \url{https://arxiv.org/abs/2504.08066}.

\bibitem[Yu et~al.(2025)Yu, Cohen, Ghazi, Huang, Kamath, Kumar, Liu, and
  Zhang]{yu2025scone}
Da~Yu, Edith Cohen, Badih Ghazi, Yangsibo Huang, Pritish Kamath, Ravi Kumar,
  Daogao Liu, and Chiyuan Zhang.
\newblock Scaling embedding layers in language models.
\newblock In D.~Belgrave, C.~Zhang, H.~Lin, R.~Pascanu, P.~Koniusz,
  M.~Ghassemi, and N.~Chen, editors, \emph{Advances in Neural Information
  Processing Systems}, volume 38, Main Conference, pages 31969--31995. Curran
  Associates, Inc., 2025.
\newblock \doi{10.52202/085713-1074}.
\newblock URL
  \url{https://proceedings.neurips.cc/paper_files/paper/2025/file/2e067924aeeb02ae9919803fd08d8b4b-Paper-Conference.pdf}.

\bibitem[Yu et~al.(2026)Yu, Yu, Zhu, Wang, Li, Peng, Zhao, Yang, He, Guo, Ren,
  Qu, Tang, and Liu]{yu2026ophis}
Shanbin Yu, Danni Yu, Baiyu Zhu, Jinxin Wang, Peihang Li, Zirui Peng, Haoran
  Zhao, Leqian Yang, Yuting He, Shaoyang Guo, Zirui Ren, Qingyu Qu, Chencheng
  Tang, and Ziming Liu.
\newblock {OPHIS}: Towards mechanistic auto-research.
\newblock MetaCircle Blog, July 2026.
\newblock URL
  \url{https://meta-circle.com/blog/ophis-a-new-paradigm-for-autoresearch}.

\bibitem[Zhang et~al.(2017)Zhang, Bengio, Hardt, Recht, and
  Vinyals]{zhang2017rethinking}
Chiyuan Zhang, Samy Bengio, Moritz Hardt, Benjamin Recht, and Oriol Vinyals.
\newblock Understanding deep learning requires rethinking generalization.
\newblock In \emph{International Conference on Learning Representations}, 2017.
\newblock URL \url{https://openreview.net/forum?id=Sy8gdB9xx}.

\bibitem[Zhou et~al.(2023)Zhou, Liu, Xu, Iyer, Sun, Mao, Ma, Efrat, Yu, Yu,
  Zhang, Ghosh, Lewis, Zettlemoyer, and Levy]{zhou2023lima}
Chunting Zhou, Pengfei Liu, Puxin Xu, Srinivasan Iyer, Jiao Sun, Yuning Mao,
  Xuezhe Ma, Avia Efrat, Ping Yu, Lili Yu, Susan Zhang, Gargi Ghosh, Mike
  Lewis, Luke Zettlemoyer, and Omer Levy.
\newblock {LIMA}: Less is more for alignment.
\newblock In \emph{Advances in Neural Information Processing Systems},
  volume~36, pages 55006--55021. Curran Associates, Inc., 2023.
\newblock URL
  \url{https://proceedings.neurips.cc/paper_files/paper/2023/hash/ac662d74829e4407ce1d126477f4a03a-Abstract-Conference.html}.

\end{thebibliography}
\bibliographystyle{plainnat}
\clearpage
\appendix
\section{Related Work}
\label{app:related}

\paragraph{Over-encoding.}
Adding $n$-gram embeddings is one way to implement over-encoding. The Over-Tokenized Transformer adds hashed embeddings of several $n$-gram orders to the input embedding and reports performance comparable to baselines about twice its size at no additional compute \citep{huang2025over}. Related methods construct and use $n$-gram representations in different ways: FastText sums character $n$-gram vectors \citep{bojanowski2017subword}; N-Grammer embeds latent bigram ids \citep{roy2022ngrammer}; SCONE learns contextual embeddings for frequent $n$-grams and keeps them off-accelerator at inference \citep{yu2025scone}; and BLT uses hashed byte $n$-gram embeddings to form patch representations \citep{pagnoni2025blt}. Engram retrieves hashed $n$-gram embeddings at several layers and gates them into the residual stream, outperforming an iso-parameter, iso-FLOPs mixture-of-experts baseline \citep{cheng2026engram}; an Engram memory with 196B parameters ships in the open-weight DeepSeek-V4.1-Flash \citep{deepseek2026v41flash}. An automated search for NanoChat training recipes also found hashed bigram and trigram embeddings injected into attention values \citep{recursive2026research}. Beyond these $n$-gram constructions, more general techniques provide ways to represent large feature sets compactly. Feature hashing maps an unbounded feature space into a fixed number of rows at the cost of collisions \citep{weinberger2009hashing}, while Hash Embeddings combine shared hashed components with learned importance weights \citep{svenstrup2017hash}. In our work, we compare three injection routes under repeated passes over a fixed corpus and characterize the resulting forking.

\paragraph{Other memory architectures.}
Product-key memories \citep{lample2019product} and memory layers at scale \citep{berges2025memory} add large trainable key--value tables with sparse access, and feed-forward layers can themselves be read as key--value memories \citep{geva2021ffn}. Non-parametric alternatives interpolate with nearest neighbors in a datastore \citep{khandelwal2019knnlm,xu2023whyknn}, attend to retrieved chunks \citep{borgeaud2021retro}, cache recent hidden states \citep{grave2016cache,wu2022memorizing}, or query unbounded $n$-gram counts \citep{liu2024infinigram}. Memory Grafting precomputes values for frequent $n$-grams with a separate model and freezes them during pretraining \citep{cheng2026memorygrafting}. An $n$-gram table differs from learned-key memories in that its routing is fixed by the input tokens rather than learned; we show that this fixed routing, which gives each context its own coordinates, is what lets repetition turn into per-context memorization (Section~\ref{sec:sharpening}).

\paragraph{Training with data replay.}
Repeated training can be useful when data are limited: up to about four epochs of repetition are nearly as useful as fresh data, with rapidly diminishing returns afterwards \citep{muennighoff2023scaling}. Multi-epoch training is also used in successful fine-tuning recipes; LIMA trains on 1{,}000 examples for 15 epochs \citep{zhou2023lima}.

Other studies document its costs. Repeating a small subset can disproportionately harm test loss and damage copying and induction-head circuits \citep{hernandez2022repeated}; duplication increases verbatim extraction, which deduplication reduces \citep{carlini2023quantifying,lee2022dedup,kandpal2022dedup}. Concurrently with our work, \citet{jha2026datascarcity} find that mixture-of-experts models overfit repeated data more than dense models and degrade at lower repetition counts.

Further work suggests that the effects of repetition depend on the training conditions and the measure of performance. Dataset size, model size, and training objective affect multi-epoch degradation, and dropout is the only common regularizer found to help substantially in the study of \citet{xue2023repeat}. Hyperfitting reports worse validation loss but better long-form generation after fitting a small set to near-zero training loss \citep{carlsson2025hyperfitting}. More generally, neural networks can fit even random labels \citep{zhang2017rethinking}; larger language models memorize faster and forget less \citep{tirumala2022memorization}, yet memorization of specific sequences is hard to predict from smaller runs \citep{biderman2023emergent}. \citet{howard2023singleexample} report abrupt decreases in training loss at epoch boundaries, sometimes accompanied by increasing validation loss, during language-model fine-tuning. They interpret these patterns in terms of rapid memorization and overconfidence. Calibration is a separate concern: neural networks can be overconfident relative to their accuracy \citep{guo2017calibration}. Architecture further changes the response to replay: in the MoE study, overfitting grows with sparsity as set by total rather than active parameters, routing stabilizes early, and expert specialization correlates with overfitting \citep{jha2026datascarcity}. Building on these observations, we investigate replay-induced train--validation divergence in greater detail and identify over-encoding as a factor that amplifies it. Through controlled interventions and context-frequency analysis, we examine how repeated training sharpens observed continuations while suppressing unseen alternatives.

\paragraph{Natural-language distributions.}
Long-tailed distributions make rare events difficult to cover in a finite sample. Memorizing rare examples can be necessary for near-optimal generalization under such distributions \citep{feldman2020memorization,feldman2020longtail}, while language models learn knowledge about rare entities poorly \citep{kandpal2023longtail}. Good--Turing estimation infers the probability of unseen outcomes from singleton counts \citep{good1953population,gale1995good}, and occupancy theory relates the decay of this missing mass to the tail of the sampling distribution \citep{gnedin2007occupancy}. Classical language models address sparse observations through smoothing: maximum-likelihood estimates assign zero probability to unobserved events, while Katz back-off, Kneser--Ney smoothing, and hierarchical Pitman--Yor models reserve probability mass for them in different ways \citep{katz1987estimation,kneser1995backing,chen1999smoothing,teh2006hierarchical}. In our work, we study these distributional properties at the level of context-specific continuations, linking poor coverage in rare contexts to forking under replay (Appendix~\ref{app:coverage}). Low-frequency masking provides a simple analogue of back-off.

\paragraph{Automated research.}
Automated research spans the evaluation of individual experimental tasks and the automation of a broader research process. MLAgentBench benchmarks language-model agents on machine-learning experimentation \citep{huang2024mlagentbench}. The AI Scientist and subsequent work extend the scope to idea generation, experiments, and paper writing \citep{lu2024scientist,yamada2025scientistv2,lu2026nature}. Another line of work focuses on improving training code under a fixed computational budget. NanoGPT provides a small language-model training implementation \citep{karpathy2022nanogpt}, while Autoresearch and Recursive's searches explore training recipes using fixed-budget metrics such as validation bits per byte \citep{karpathy2026autoresearch,recursive2026research}. These searches offer a setting in which changes to an architecture or training procedure can be evaluated through repeated experiments. OPHIS guides such searches with measurements of internal training dynamics \citep{yu2026ophis}, bringing the behavior of model components into the investigation alongside the target metric. Forking was first observed during an OPHIS-guided architecture search on NanoGPT (Appendix~\ref{app:discovery}). Our work illustrates a by-product of automated search: a component selected for a better target metric under a single-pass budget can fail under a different data regime. Following this observation, we examine the failure under replay and use interventions to investigate its mechanism.

\section{Original Task and OPHIS Experiments}
\label{app:discovery}

\subsection{The original task and discovery setting}
\label{app:original-setting}

NanoGPT RSI searches for changes that reduce validation bits per byte (BPB). An agent modifies \code{train.py} within a fixed compute budget, evaluates the result, and continues to revise the architecture and training strategy. The original automated search used a training budget of approximately five minutes per experiment \citep{karpathy2026autoresearch,recursive2026research}. The small-data experiment in which forking was discovered instead used 2000 optimizer steps to inspect a longer trajectory. Forking was an unexpected outcome of this process.

The text came from the Hugging Face dataset \href{https://huggingface.co/datasets/karpathy/climbmix-400b-shuffle}{\code{karpathy/climbmix-400b-shuffle}}, using shards 1--10 and 6542. In the discovery-stage split, shards 1--3 were used for training, and shards 4--10 and 6542 for validation. The loader read and packed text in a fixed order, then restarted after traversing the training shards. Three shards did not provide enough unique data for 2000 steps, so the run necessarily entered later epochs. Training and validation losses initially decreased together. When text began to repeat, training loss continued to fall while validation loss flattened or increased, producing a rapid divergence near the epoch boundary.

The agent-added n-gram module used independent learnable lookup tables at different layers. Each bigram or trigram context was mapped through two hashes. The two retrieved half-width vectors were concatenated, then injected into the attention value branch through an input-dependent sigmoid gate:
\begin{equation}
 v_t^{(\ell)}\leftarrow v_t^{(\ell)}
 +g_{2,t}^{(\ell)}\odot e_{2,t}^{(\ell)}
 +g_{3,t}^{(\ell)}\odot e_{3,t}^{(\ell)}.
 \label{eq:original-injection}
\end{equation}
Here $e_{n,t}^{(\ell)}$ is the concatenated lookup result; the gate is computed from the corresponding channels of that layer's input. Hash constants and table parameters differ across layers. The retrieved vectors enter the value branch, rather than being added directly to the token embedding.

Table~\ref{tab:discovery-config} records the frozen configuration used in the observable-localization experiments below. It differs from the later controlled setting in Section~\ref{sec:setting}.

\begin{table}[!htbp]
\centering\small
\caption{Configuration for discovery-stage observable screening. The architecture and training settings were determined by the search agent and held fixed during screening. Layer indices start at one.}
\label{tab:discovery-config}
\begin{tabular}{@{}p{0.21\linewidth}p{0.74\linewidth}@{}}
\toprule
Component & Configuration\\
\midrule
Backbone & 8 layers; hidden dimension 768; 6 attention heads; head dimension 128.\\
Input and batch & Sequence length 2048; 72 sequences or 147456 tokens per update; gradient accumulation 1.\\
Tokenizer & Shared tokenizer; vocabulary 8192; built from the first ten shards.\\
N-gram tables & Two tables per order and injection layer; each has 524288 rows of dimension 384. Bigram at layers 2, 4, 6, 8; trigram at layers 2, 6, 8.\\
Optimizers & Muon for matrix parameters, base learning rate 0.04; grouped AdamW for token/value embeddings, output and scalar parameters; RMSProp for bigram/trigram tables, learning rate 0.6, second-moment decay 0.999, no weight decay.\\
Schedule & Archived baseline schedule: Muon and ordinary AdamW warmdown begins at 5\% and 35\% of the step budget, respectively; final multiplier 0.05. N-gram table learning rates remain constant.\\
Logging & Four runs, each 1000 steps, seed 42. Validation uses four batches every five steps. Optimizer observables are normally logged every five steps and at every step in the 320--380 window, alongside parameter and layer statistics.\\
\bottomrule
\end{tabular}
\end{table}

\subsection{Localizing the n-gram pathway across controlled replay schedules}
\label{app:screening}

We sought observables whose changes consistently tracked forking, so that their parameter-group attribution could help locate the responsible component. OPHIS recorded too many observables for manual inspection of every curve, making an automated screening procedure necessary. Starting from the single-shard replay experiment, we added controls that changed when and how training data were reused. We then defined screening rules from the expected timing and shape of forking in these experiments. Only observables that passed every rule were retained for manual review. Finally, we disabled the implicated component to test its contribution to forking.

\subsubsection{Four experiments used to screen observables}

We varied when training data were first reused and how reuse was organized, while fixing the model, optimizer configuration, tokenizer, and step budget. The four experiments distinguish ordinary replay, delayed replay, random repetition without a full-pass transition, and an imposed replay cycle.

\paragraph{E1 (TR1): sequential replay of one shard.}
The loader repeatedly traversed shard 1 in a fixed order. The second epoch started at step 338, followed by pronounced forking; the third started at step 687, with a further increase in the gap. This run supplied the boundary event to be localized.

\paragraph{E2 (TR2): sequential replay of two shards.}
The loader used shards 1 and 2 before restarting. At approximately step 350, it was still in the first traversal, and the training and validation curves remained close. The second epoch started at step 683, and the divergence was delayed accordingly. This control distinguishes a response to reuse from a change tied only to the absolute training step.

\paragraph{E3: random sampling with replacement.}
Documents were sampled uniformly with replacement from shard 1, then tokenized and packed into batches. Repeated documents could occur from the beginning of training. A gap developed gradually, but there was no full-pass replay transition near step 350. This supplied a control for the local boundary event.

\paragraph{E4: 50-new / 50-replay.}
The loader read 50 consecutive new batches from the shard-1 stream, then replayed those cached batches in the same order. Training loss fell and validation loss rose during replay; the curves moved closer when new data resumed. The resulting cycle of approximately 100 steps provided a test of whether an observable followed the replay state. Near the end of a base-stream epoch, a short block was replayed only for its actual cached length, so the local period could be shorter than 100 steps.

Experiments E1, E3, and E4 used validation shards 2--10 and 6542; E2 used shards 3--10 and 6542. These validation sets were therefore not identical. Screening compared the temporal responses of internal observables; loss curves identified the event within each run. In this appendix, a step counts completed optimizer updates. The centers 350 and 700 refer to screening windows, while actual epoch boundaries come from loader logs.

\begin{figure}[!htbp]
\centering
\includegraphics[width=\linewidth]{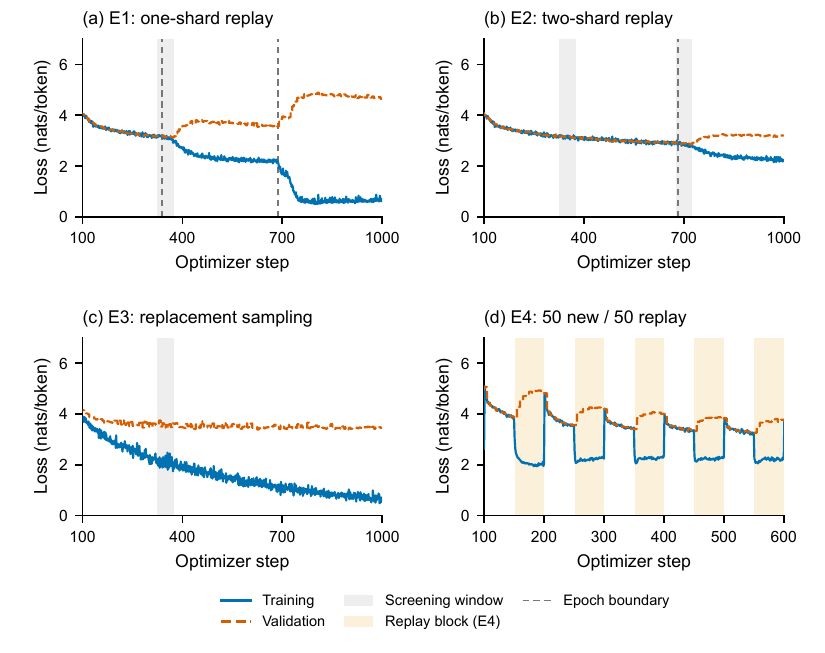}
\caption{Four data-reuse controls (E1--E4) in the discovery setting, each with seed 42 and 1000 updates. Blue solid and orange dashed curves show training and validation cross-entropy. Gray bands mark $350\pm25$ and, for E2, $700\pm25$; vertical dashed lines mark logged epoch boundaries. E4 shows steps 100--600, with replay blocks shaded orange; the other panels show steps 100--1000. Training losses are unsmoothed. Validation uses four successive held-out batches every five steps, with the latest value carried forward for display.}
\label{fig:four-controls}
\end{figure}

\subsubsection{Five observable-screening rules}

OPHIS records statistics of parameters, gradients, optimizer states, and update vectors, including their directional relationships \citep{yu2026ophis}. Candidates were scalar time series shared by the four runs. We excluded bookkeeping quantities such as data progress, learning-rate multipliers, throughput, and parameter counts, as well as some auxiliary fields used to construct other statistics. Window calculations used finite recorded values without interpolation.

Let $\mu(X)$ and $\sigma(X)$ denote the mean and population standard deviation of the recorded values in $X$, let $X\sqcup Y$ concatenate both collections, and set $\varepsilon=10^{-12}$. To compare windows or runs with different scales, we used
\begin{equation}
 D(X,Y)=\frac{|\mu(X)-\mu(Y)|}
 {\max\{\sigma(X\sqcup Y),\,0.05|\mu(X\sqcup Y)|,\,\varepsilon\}}.
\end{equation}
For a local event compared with a neighboring baseline $B$, we used
\begin{equation}
 H(X,B)=\frac{|\mu(X)-\mu(B)|}
 {\max\{\sigma(X\sqcup B),\,0.05|\mu(B)|,\,\varepsilon\}}.
\end{equation}
The windows were $W=[325,375]$, $L=[255,315]$, $R=[385,445]$, and the later E2 window $W'=[675,725]$. Write $X_r(W)$ for the values recorded in window $W$ of experiment $r\in\{\mathrm{E1},\mathrm{E2},\mathrm{E3},\mathrm{E4}\}$.

\paragraph{C1: a within-run change across replay states.}
Compare E2 near steps 350 and 700:
\begin{equation}
 C_1=D\bigl(X_{\mathrm{E2}}(W),X_{\mathrm{E2}}(W')\bigr)\geq0.5.
\end{equation}
The early window is within the first traversal; the later one covers the onset of replay. This rule retains curves that differ across these states.

\paragraph{C2: a cross-run difference at the same training time.}
Compare E1 and E2 near step 350:
\begin{equation}
 C_2=D\bigl(X_{\mathrm{E1}}(W),X_{\mathrm{E2}}(W)\bigr)\geq0.5.
\end{equation}
The runs have the same elapsed optimizer steps and learning-rate schedule, but only E1 has entered replay.

\paragraph{C3: a boundary event stronger than random repetition.}
For each run, define
\begin{equation}
 E_r=H\bigl(X_r(W),X_r(L)\sqcup X_r(R)\bigr).
\end{equation}
Require
\begin{equation}
 C_3=\max(0,E_{\mathrm{E1}}-E_{\mathrm{E3}})\geq0.5.
\end{equation}
This retains candidates whose local E1 event exceeds the change at the same time under random repetition in E3.

\paragraph{C4: agreement with the imposed replay period.}
Within the first 700 steps of E4, pair all recorded values separated by exactly 100 steps. Let $U,V$ be the paired columns and $\rho_{100}$ their Pearson correlation. Define
\begin{equation}
 a=\frac{\min\{\sigma(U),\sigma(V)\}}
         {\max\{\sigma(U\sqcup V),\varepsilon\}},\qquad
 C_4=\max(0,\rho_{100})\min(1,a)\geq0.15.
\end{equation}
Only pairs with $t\geq1$ and $t+100\leq700$ are used. The score is zero when fewer than ten pairs are available.

\paragraph{C5: a center window distinct from both neighbors.}
Require the E1 center to differ from both the preceding and following windows:
\begin{equation}
 C_5=\min\left\{
 \begin{aligned}
 &H\bigl(X_{\mathrm{E1}}(W),X_{\mathrm{E1}}(L)\bigr),\\
 &H\bigl(X_{\mathrm{E1}}(W),X_{\mathrm{E1}}(R)\bigr)
 \end{aligned}\right\}\geq0.5.
\end{equation}
Taking the smaller deviation rejects curves that differ on only one side.

A candidate had to pass all five thresholds. For manual-review ordering, the script saturated each score as $q_k=C_k/(1+C_k)$ and ranked candidates by
\begin{equation}
 S=100\left[0.7\min_k q_k+0.3\,\operatorname{mean}_k q_k\right].
\end{equation}
The windows and thresholds were exploratory heuristics. These scores narrowed the set for inspection; they were not significance tests or causal evidence.

\subsubsection{Manual inspection and component removal}

\begin{figure}[!htbp]
\centering
\includegraphics[width=\linewidth]{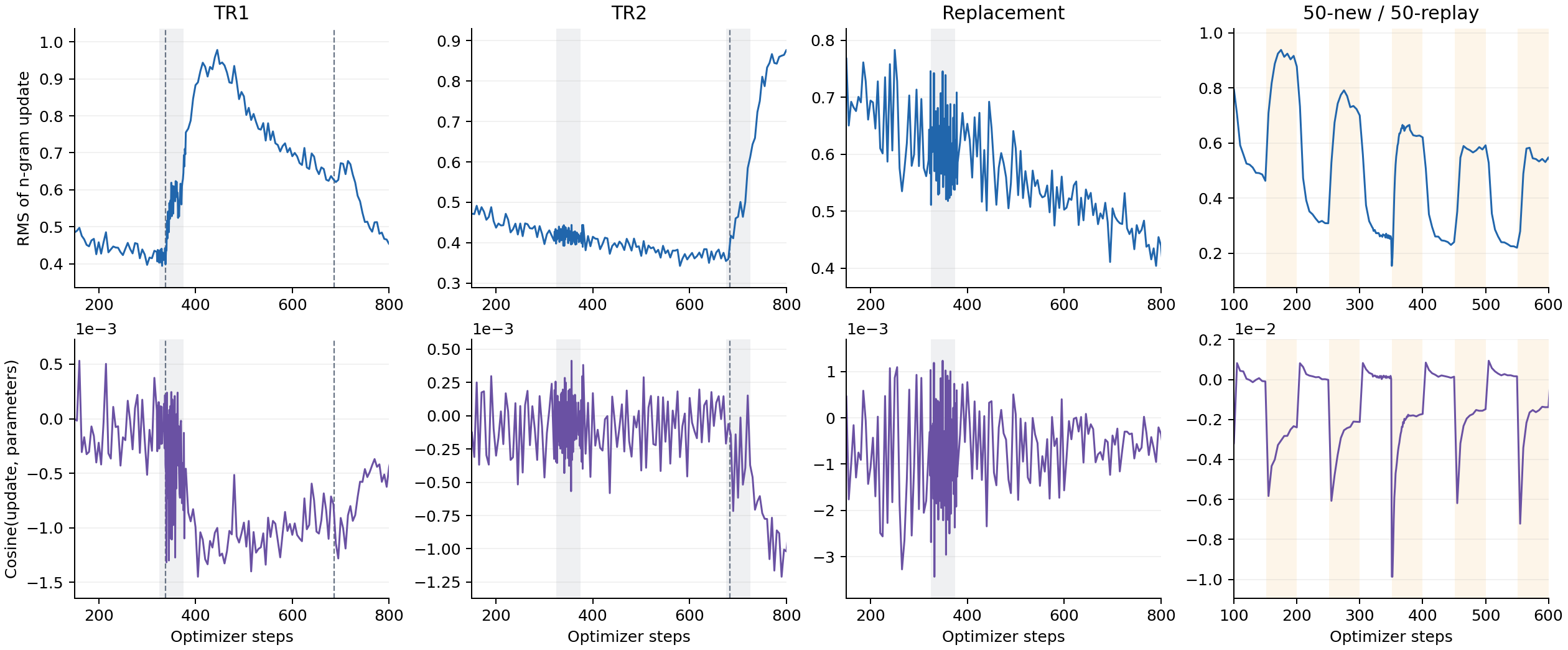}
\caption{Two representative n-gram observables. The top row shows the RMS magnitude of the RMSProp update vector; the bottom row shows its cosine similarity with the parameter vector. Columns correspond to E1--E4 from left to right. Shading and boundaries follow Figure~\ref{fig:four-controls}. Each panel uses its own visible vertical range. Unsmoothed curves are used to compare event timing and shape.}
\label{fig:ngram-observables}
\end{figure}

Curves that were identical across the four runs, or differed only by a fixed scale factor, were merged. Representative statistics were chosen for clear parameter-group attribution and interpretability. Manual inspection of the remaining small set highlighted the n-gram table's update magnitude and update--parameter direction relationship. The update magnitude rose after replay began in E1, shifted later in E2, and followed the alternation between new and replayed data in E4 (Figure~\ref{fig:ngram-observables}). Code inspection confirmed that this logged parameter group consisted of the bigram and trigram value-embedding lookup tables. The n-gram pathway therefore became the main candidate for component intervention.

\begin{figure}[!htbp]
\centering
\includegraphics[width=\linewidth]{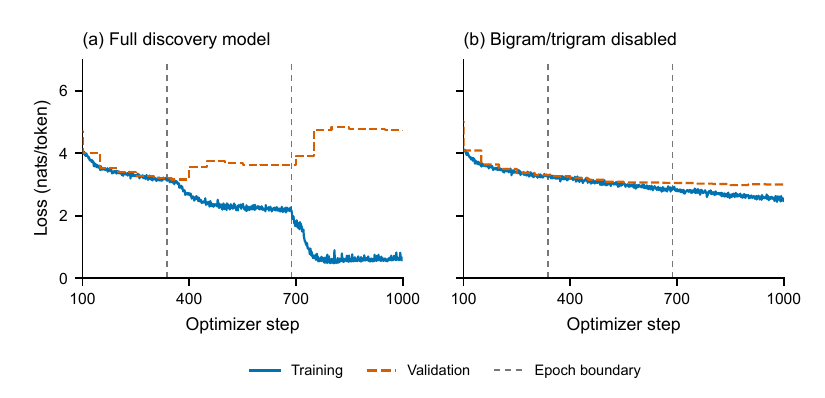}
\caption{Component removal in the discovery setting. Left: the full model. Right: bigram and trigram value pathways disabled, with unigram value embeddings retained. Both runs use shard-1 sequential replay, seed 42, and 1000 updates; the axes are shared. Blue solid curves show raw training cross-entropy. Orange dashed curves show validation cross-entropy on four successive held-out batches every 50 steps, carried forward between evaluations. Vertical dashed lines mark replay boundaries at steps 338 and 687.}
\label{fig:discovery-removal}
\end{figure}

We then trained the full model and a model with the bigram/trigram value pathways disabled under the same single-shard sequential-replay setting, 1000-step budget, and seed. Unigram value embeddings were retained. The full model forked sharply after replay began. Disabling the two higher-order pathways removed this pattern, and validation loss continued to improve in later epochs (Figure~\ref{fig:discovery-removal}). Final validation BPB was 1.6769 for the full model and 1.0768 for the intervention. This supports a causal role for the n-gram pathway in this configuration.

\section{Minimal Setting, Parameter Choices, and Measurement Conventions}
\label{app:setting}

This appendix specifies the minimal setting used to reproduce and study forking after the discovery-stage localization in Appendix~\ref{app:discovery}. We first describe the fixed architecture and data protocol, then document the optimization choices and their supporting experiments, followed by measurement and numerical conventions.

\subsection{Architecture and fixed experimental settings}
\label{app:minimal-architecture}

We use a vanilla NanoGPT Transformer \citep{karpathy2022nanogpt} with full causal attention, learned absolute position embeddings, pre-LayerNorm blocks, and tied token and output embeddings. Table~\ref{tab:minimal-config} records the default configuration. The model retains the discovery setting's depth, hidden width, head count, vocabulary size, sequence length, and batch size, while replacing its specialized architecture with standard Transformer blocks. These dimensions are held fixed to keep the comparisons controlled; they were not selected by an architecture search for the largest gap.

\begin{table}[!htbp]
\centering\small
\caption{Default architecture and data settings for the minimal experiments. A listed quantity changes only in experiments that explicitly vary it. Optimization settings and their selection evidence are given in Appendix~\ref{app:parameter-selection}.}
\label{tab:minimal-config}
\begin{tabular}{@{}p{0.23\linewidth}p{0.72\linewidth}@{}}
\toprule
Component & Configuration\\
\midrule
Attention backbone & 8 layers; hidden width 768; 6 heads of dimension 128; full causal attention in every layer.\\
Feed-forward blocks & Two linear layers with a $4\times$ expansion to width 3072 and a GELU activation.\\
Normalization and positions & Pre-LayerNorm in each block and a final LayerNorm, $\epsilon=10^{-5}$; learned absolute position embeddings.\\
Embeddings and regularization & Vocabulary 8192; tied token/output weights; dropout 0. Attention and MLP linear layers and LayerNorm include biases; the output head does not.\\
N-gram memory & One $2^{20}\times768$ hash table per order, bigram and trigram; full-width lookup vectors added at the input. No extra unigram or fourgram memory, gates, or multi-hash concatenation.\\
Sequence and batch & 2048 prediction positions per sequence; 72 sequences per update; 147,456 prediction tokens; one device batch per update.\\
Training data and budget & Shard 1 in a fixed replay order; 337 updates per pass. The initial diagnostic budget is 1000 updates; extended comparisons commonly use 2000.\\
Validation pool & Eligible shards 2--10 and 6542, disjoint from training. The first four batches of shard 2 form the fixed pool: 589,824 prediction positions.\\
Seeds and evaluation & Default model and data seed 42; seeds 43 and 44 where reported. Main loss and frequency measurements every 10 steps unless stated otherwise.\\
\bottomrule
\end{tabular}
\end{table}

\paragraph{Why simplify the memory pathway?}
The discovery model distributes gated, multi-hash lookup tables across attention value branches. The minimal model instead adds one full-width vector per n-gram order directly to the token representation. This makes the memory pathway easy to remove or change while keeping the backbone fixed. Bigram and trigram tables retain the two components identified in Appendix~\ref{app:screening}; separate experiments disable either or both. The input route is the default, with value and attention-output injection tested as alternatives in Section~\ref{sec:setting}. The default $2^{20}$ rows per table provides a common reference for comparisons; the capacity sweep varies this quantity explicitly. Here \emph{clean table} denotes the single-table implementation, not a guarantee of collision-free lookup.

\paragraph{Why fix the data and replay budget?}
A single shard permits several complete passes within a short run, making the onset and subsequent growth of forking observable. Keeping the tokenizer, sequence length, and batch size fixed preserves the token and context definitions and the number of prediction tokens per update. The 1000-step screen spans nearly three passes; 2000-step runs examine continued replay. Reusing the same held-out batches keeps validation-sample changes out of the measured trajectory. Seed 42 is the default reference; additional seeds quantify training-seed variation only where they were actually run. The model dimensions, batch size, table size, and default seed are fixed experimental choices, rather than claims of individually optimal hyperparameters.

\subsection{Selecting the controlled optimization setting}
\label{app:parameter-selection}

After localizing the n-gram pathway, we selected the controlled optimization setting in the vanilla architecture of Section~\ref{sec:setting}. These screens targeted a clear replay gap within a short budget; they were not validation-loss optimization. The screens use the protocol in Table~\ref{tab:minimal-config}: input injection, clean $2^{20}$-row bigram and trigram tables, shard-1 replay, fixed validation, BF16, and seed 42. The backbone uses AdamW $(\beta_1,\beta_2)=(0.8,0.95)$ with weight decay $0.1$; tables use bias-corrected RMSProp $(0,0.99)$ with no weight decay. Both learning rates warm up for 100 steps from $0.25\times$ to their maxima and then remain constant. The selected maxima are $\eta_B=6\times10^{-4}$ and $\eta_T=128\eta_B=0.0768$.

\begin{figure}[!htbp]
\centering
\includegraphics[width=\linewidth]{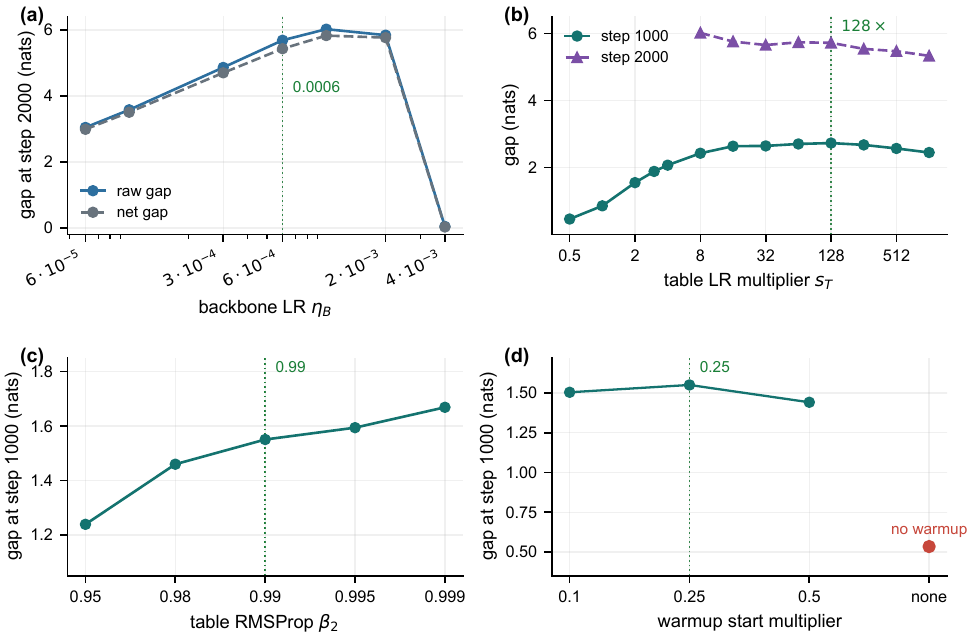}
\caption{Selection screens for the controlled optimization setting. (a) Backbone-rate sensitivity with the absolute table rate fixed at $0.0768$; solid and dashed curves show raw and matched no-gram-subtracted gaps at step 2000. (b) Table-rate multiplier scan at steps 1000 and 2000. (c) Corrected RMSProp $\beta_2$ sweep. (d) Warmup-start sweep; ``none'' removes warmup. Green vertical dotted lines mark the values adopted for the controlled setting: backbone learning rate $0.0006$, table-rate multiplier $128\times$, RMSProp $\beta_2=0.99$, and warmup-start multiplier $0.25$.}
\label{fig:parameter-selection}
\end{figure}

The initial backbone screen compared $4\times10^{-4}$, $6\times10^{-4}$, and $4\times10^{-3}$ at a fixed $2\times$ table multiplier; their step-1000 gaps were $1.187$, $1.534$, and $0.060$. Because the absolute table rate changed too, that screen did not isolate the backbone rate. Panel (a) is a later check holding the absolute table rate fixed; $6\times10^{-4}$ lies in a broad large-gap regime, whereas $4\times10^{-3}$ underfits. The table scan reaches $G_{1000}=2.732$ at $128\times$, within a broad high-gap range. At $2\times$, the corrected $\beta_2$ sweep spans $0.95$--$0.999$; $0.99$ is the standard value, not the gap maximum. The $0.25$ warmup start is near neighboring starts, whereas removing warmup changes the trajectory substantially.

\paragraph{Scope and provenance.}
Screens use seed 42 unless stated otherwise. Discovery-stage configurations and incomplete runs are excluded.

\subsection{Loss and frequency measurements}

The discovery-stage configuration in Appendix~\ref{app:original-setting} and the controlled setting in Section~\ref{sec:setting} should be distinguished. They use different optimizers and injection implementations. Frequencies $F(i)$ refer to counts in a single unrepeated training corpus, even when that corpus is consumed many times.

For online curves, the training loss is obtained from the training forward pass before the corresponding update, while validation can be evaluated after that update. A fixed-probe gap instead compares fixed training and validation batches evaluated at the same checkpoint:
\begin{equation}
 G_{\mathrm{probe}}(t)
 =L_{\mathrm{val,probe}}(t)-L_{\mathrm{train,probe}}(t).
\end{equation}
Where a matched no-gram control is subtracted, we explicitly use
\begin{equation}
 G_{\mathrm{net}}(t)
 =G_{\mathrm{ngram}}(t)-G_{\mathrm{nogram}}(t).
\end{equation}
Cross-entropy values are in nats unless stated otherwise. BPB is the original search metric and should not be numerically equated with token cross-entropy.

\subsection{Numerical precision and implementation}

The main setting uses BF16 autocast for forward computation. Model weights, n-gram table parameters, and optimizer states remain in FP32. Cross-entropy is computed from FP32 logits; per-token losses are returned in FP32 and accumulated in FP64 for frequency-conditioned diagnostic statistics. The current diagnostic forwards use the same BF16 convention as the main evaluation. Earlier diagnostic runs that used FP32 forwards retain their recorded precision setting. Asynchronous aggregation changes execution, not the metric definitions.

FP8 was not used as the baseline. In a setting-validation test, simply selecting \code{float8_e4m3fn} in CUDA autocast did not work on the current eager \code{nn.Linear}/\code{addmm} path. FP8 training requires a dedicated implementation, such as scaled matrix multiplication or Transformer Engine, together with explicit scaling policies for activations, weights, and gradients. Precision boundaries would also have to be specified separately for lookup tables, attention, MLPs, logit projection, table optimization, and diagnostics.

A historical matched-hyperparameter comparison on an H200 found BF16 approximately 5.3 times faster per step than FP32. At 1000 steps, the reported raw gaps were 0.825 for BF16 and 0.858 for FP32, with similar trajectories. This check supported the conclusion that the numerical format did not remove forking in that setting; it is not a substitute for the evidence from the current main runs. The default convention remains BF16 forward computation, FP32 weights and optimizer states, and no \code{torch.compile}. Any future FP8 comparison should report stability, loss, gap, margins, throughput, and memory use separately.

\section{Full Intervention and Replay Curves}
\label{app:interventions}

\begin{figure}[!htbp]
\centering
\includegraphics[width=\linewidth]{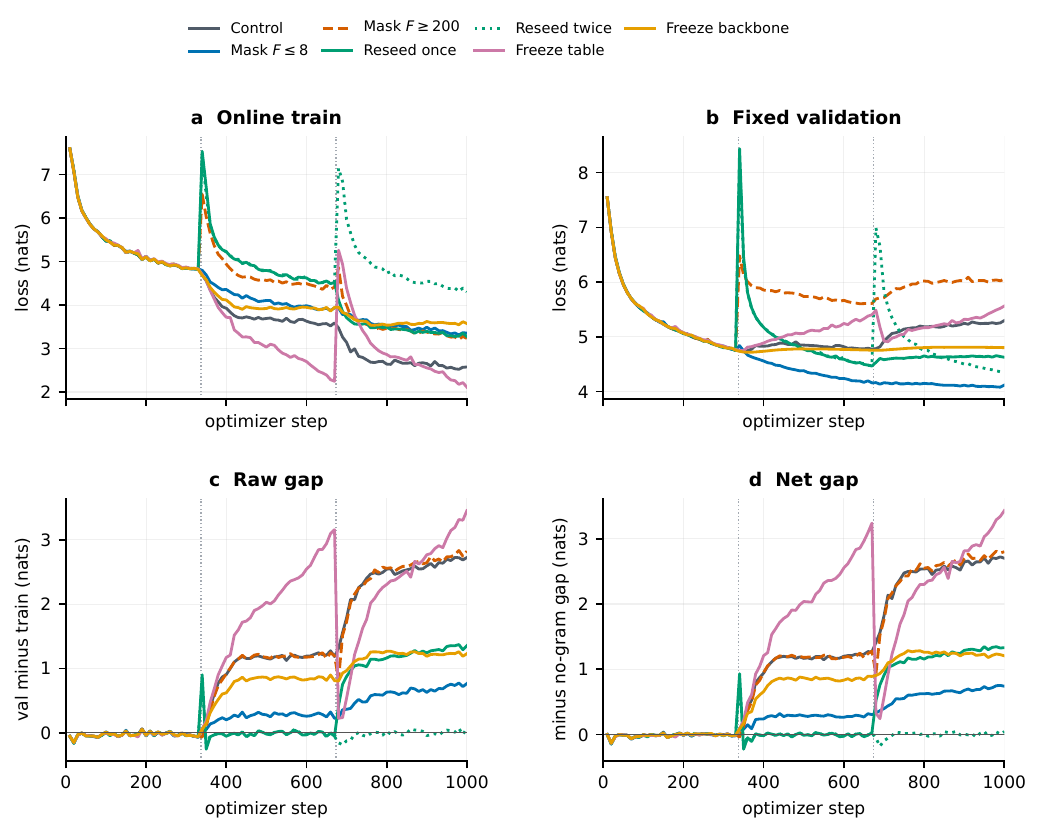}
\caption{Training, validation, and gap trajectories for the component interventions (seed 42, logged every 10 steps). Net gap subtracts the no-gram model at matching steps; dotted lines mark the replay boundaries at steps 337 and 674. The hash-reseeding conditions change the context-to-row mapping while retaining table weights and optimizer state, either once at step 338 or at steps 338 and 675. The twice-reseeded arm ends with raw gap 0.069.}
\label{fig:causal-full}
\end{figure}

Figure~\ref{fig:causal-full} and Table~\ref{tab:causal} summarize the main intervention evidence. Each intervention addresses a different part of the proposed pathway: context-dependent reads, stable hash assignment, table writing, or backbone adaptation. Interpretations are restricted to the tested thresholds and measurement windows.

Section~\ref{sec:interventions} reports low-frequency masking and table freezing; this appendix gives the full set. Frequency masking sets the retrieved embedding to zero for contexts in a chosen count range and tests which contexts' reads carry the gap. Hash reseeding replaces the context-to-row mapping while keeping the table weights and optimizer state, so rows written in earlier passes no longer serve the same contexts; it tests whether the gap needs a stable key for each context across passes. Freezing the table or the backbone stops one of the two update paths and tests which one drives the growth after the first pass.

All runs in Figure~\ref{fig:causal-full} use the both-table input-injection model with table learning rate $128\times$ the backbone rate, seed 42, and 1000 updates on the fixed single-shard corpus (337 updates per pass). Raw gap is the fixed-validation loss minus the online loss of the current training batch; training loss is recorded before the update and validation loss after it. Net gap subtracts the gap of a separately trained no-gram model at the same step. Every intervention takes effect at step 338, the first update after the replay boundary at step 337; the twice-reseeded arm is reseeded again at step 675, after the second boundary. At step 1000 the control has raw gap 2.724. Reseeding at both boundaries reduces it to 0.069: training loss rises to 4.304 and validation loss falls to 4.373, which is still above the 4.119 reached by $F\leq8$ masking. A single reseed at step 338 leaves 1.354. After a transient spike at the reseed, its gap stays near zero for the rest of the second pass and grows again in the third, when contexts return to rows written under the new mapping. Freezing the backbone gives 1.230. Freezing the table gives 3.452, above the control, and its validation loss is also higher (5.563 versus 5.298).

\begin{table}[!htbp]
\centering\small
\caption{Intervention summary across comparisons, not a list of endpoints for Figure~\ref{fig:causal-full}. Raw gaps and relative reductions have different denominators. All runs use input injection, table LR $128\times$, and seed 42; each row states its budget.}
\label{tab:causal}
\begin{tabular}{@{}p{0.18\linewidth}p{0.34\linewidth}p{0.42\linewidth}@{}}
\toprule
Condition & Intervention & Reported observation\\
\midrule
No-gram & Remove the n-gram module & Raw gap 0.233 at step 2000; ordinary train--validation differences remain.\\
N-gram & Input injection; table LR $128\times$ & Raw gap 5.676 at step 2000, with replay-related steps.\\
Low-frequency mask & Mask low-count reads in training and evaluation & $F\leq8$ removes about 72\% of net gap at step 1000; higher thresholds further reduce it (raw gap 0.101 at $F\leq200$).\\
High-frequency mask & Mask high-count reads & $F\geq200$ leaves the step-1000 raw gap at 2.824 (control 2.724), a much smaller change than low-frequency masking.\\
Hash reseed & Change context-to-row mapping at each replay boundary; retain parameters and optimizer & Raw gap 0.069 at step 1000 when reseeded at steps 338 and 675 (1.354 with one reseed); substantial suppression of replay accumulation.\\
Freeze table & Stop table updates at step 675 & 94\% of the control gap remains at step 2022 (5.386 versus 5.733) and 72\% after ten passes; the gap keeps growing, more slowly than in control. Frozen at step 338, the step-1000 gap is 3.452.\\
Freeze backbone & Stop backbone updates at step 338 & Smaller second-pass gap (0.81 versus 1.19 at step 670); little further growth in epoch 3 (1.230 at step 1000).\\
\bottomrule
\end{tabular}
\end{table}

\begin{figure}[!htbp]
\centering
\includegraphics[width=\linewidth]{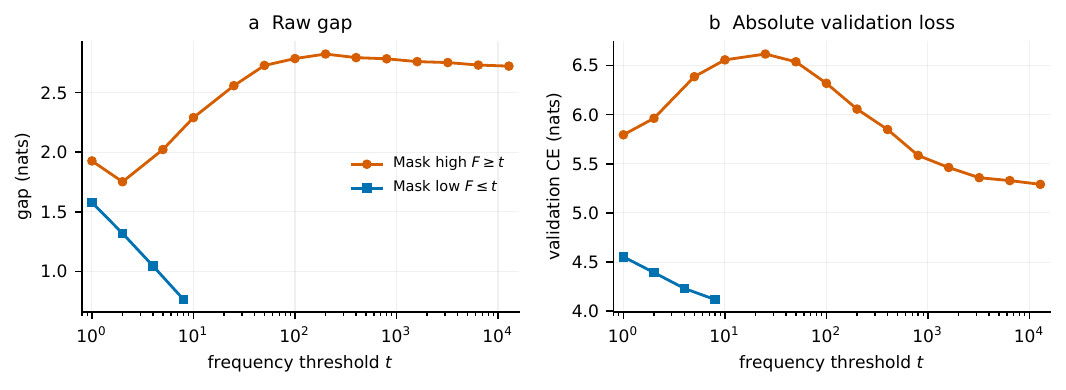}
\caption{Frequency-threshold scans at step 1000 (seed 42): (a) raw gap and (b) fixed-validation loss. High masking uses $F\geq t$ and low masking $F\leq t$. Each point is one run; lines are visual guides.}
\label{fig:causal-thresholds}
\end{figure}

Figure~\ref{fig:causal-thresholds} varies the masking threshold. High-frequency masking removes the reads of contexts with $F\geq t$ for 14 thresholds from $t=1$ to $t=12800$; low-frequency masking removes reads with $F\leq t$ for $t\in\{1,2,4,8\}$. Both start at step 338 and apply in training and evaluation. High-frequency masking never masks novel contexts ($F=0$), whereas low-frequency masking always does. At step 1000, $F\leq8$ gives raw gap 0.765 and validation loss 4.119. For $t\geq50$, high-frequency masking leaves the raw gap between 2.72 and 2.82, close to the control, and validation loss between 5.29 and 6.54, against 5.30 for the control; $F\geq200$ gives 2.824 and 6.056. Masking every seen context ($F\geq1$) lowers the raw gap to 1.927, but validation loss remains 5.794.

\begin{figure}[!htbp]
\centering
\includegraphics[width=\linewidth]{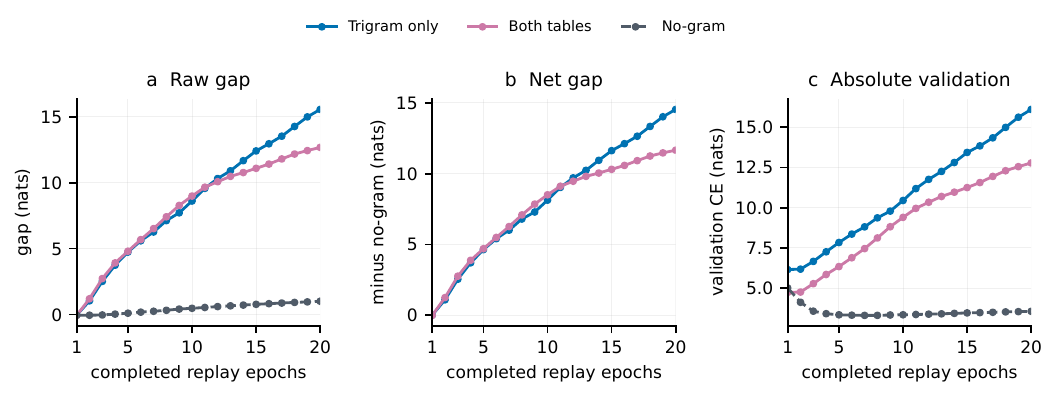}
\caption{Epoch-end raw gap, net gap, and absolute fixed-validation loss for trigram-only, both-table, and no-gram models (20 epochs, seed 42). Net gap subtracts the no-gram model at each boundary.}
\label{fig:long-replay}
\end{figure}

Figure~\ref{fig:long-replay} extends replay to 20 passes (6740 updates) for the trigram-only, both-table, and no-gram models, with validation measured at every boundary. At epoch 20 the raw gaps are 15.552, 12.674, and 1.012, and the validation losses are 16.103, 12.779, and 3.564. The no-gram model also overfits slowly: its validation loss is lowest at epoch 7 (3.31) and rises afterwards. The trigram-only net gap grows approximately linearly over passes 5--20, with a slope of 0.664 per pass and $R^2=0.998$. The same epoch-end gaps appear in Figure~\ref{fig:staircase}b. In the 2000-step n-gram run of Table~\ref{tab:causal}, boundary jumps account for approximately 28\% of the gap accumulated over passes 2--6, and their share decreases over passes. This motivates a dynamical account that includes within-pass changes, rather than attributing all damage to an instantaneous epoch transition. A finite-window linear fit does not establish unbounded long-run growth.

\begin{figure}[!htbp]
\centering
\includegraphics[width=\linewidth]{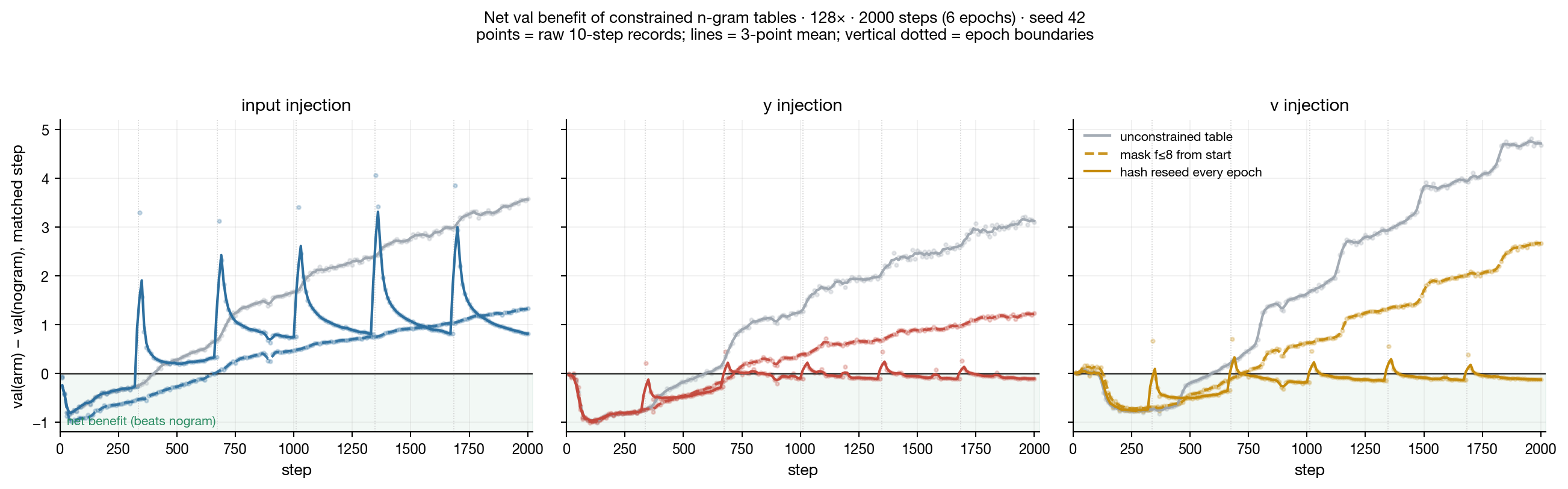}
\caption{Validation-loss differences relative to the no-gram baseline for unconstrained tables, masking of $F\leq8$ reads from the start, and hash reseeding at every replay boundary (both tables, 2000 updates, seed 42). These comparisons delimit the benefit of memory in the measured replay regime; they do not establish that n-gram memory should only be used as a short-term cache.}
\label{fig:net-benefit}
\end{figure}

A reduced gap must be distinguished from improved validation performance: increasing training loss can shrink the difference without repairing generalization. Likewise, a gap that survives table freezing does not imply that all relevant memory is stored in the table. In the 2000-step comparison of Table~\ref{tab:causal}, validation loss is approximately 6.95 with n-gram memory and 3.34 without it. Figure~\ref{fig:net-benefit} compares absolute validation changes relative to the no-gram baseline. In this regime, the additional memory does not provide a net generalization benefit.

Figure~\ref{fig:net-benefit} also tests two constraints over 2000 updates (six passes) for each injection route: masking reads with $F\leq8$ from the first step, and reseeding the hash at every replay boundary. With input injection neither constraint reaches the no-gram validation loss. At step 2000 the excess is 3.58 nats without a constraint, 1.33 with masking, and 0.81 with reseeding, and each reseed causes a transient validation spike of roughly 2--3 nats. With value or attention-output injection, reseeding ends slightly below the no-gram model (by 0.13 and 0.11 nats), while masking stays above it by 2.67 and 1.23 nats.

Low-frequency masking and reseeding are mechanism-motivated interventions, but their practical value also depends on preserved training benefits and additional cost. The choice between refreshing memory, retaining it, or protecting selected keys requires evidence under the intended data coverage and training history.

\section{Mathematical Derivations and Model Tests}
\label{app:math}

\subsection{Conditional continuation tails and the novel-continuation rate}
For a context $c_i$ with $F_i=F$, let $y_{i,(r)}$ denote the $r$-th most likely continuation and $p_{i,(r)}$ its true conditional probability. Model the observed continuations $Y_{i,1},\ldots,Y_{i,F}$ as independent draws, with counts $n_{i,(r)}(F)=\#\{k\in\{1,\ldots,F\}:Y_{i,k}=y_{i,(r)}\}$. The novel-continuation rate $f_i(F)$ and its expectation under this sampling model are
\begin{equation}
f_i(F)=\sum_r p_{i,(r)}
\mathbf{1}\left[n_{i,(r)}(F)=0\right],
\quad
\mathbb{E}[f_i(F)]=\sum_r p_{i,(r)}(1-p_{i,(r)})^F.
\end{equation}

Assume a conditional tail $p_{i,(r)}\sim A_i r^{-s}$ with $s>1$. For large $F$ within the tail's scaling range, before finite-vocabulary effects become significant, the dominant probabilities are of order $1/F$. Approximating $(1-p_{i,(r)})^F$ by $e^{-Fp_{i,(r)}}$ and the rank sum by an integral then gives
\citep{gnedin2007occupancy}
\begin{equation}
\begin{aligned}
\mathbb{E}[f_i(F)]
&\sim
\int_1^\infty A_i r^{-s}e^{-F A_i r^{-s}}\,\mathrm{d}r\\
&\sim
\frac{A_i^{1/s}}{s}
\Gamma\left(1-\frac{1}{s}\right)
F^{-(1-1/s)}\\
&\propto F^{-\alpha},
\quad
\alpha=1-\frac{1}{s}.
\end{aligned}
\label{eq:occupancy}
\end{equation}

This provides a conditional explanation of the empirical scaling in Section~\ref{sec:coverage}. Finite vocabulary limits the power-law regime to an intermediate range. Heterogeneity across contexts may alter the observed effective exponent.

\subsection{Crowding-out within a context subspace}
\label{app:crowding-derivation}

Fix a context $c_i$, and let $a_v$ be the logit of a continuation $v$. Partition the vocabulary into the continuations of $c_i$ seen in training ($\mathcal{S}$), those that only occur in validation ($\mathcal{V}$), and the rest. With probability $p_v=\softmax{a_v}$, one can idealize the per-epoch gradient update on a training target $C\in\mathcal{S}$ by
\begin{equation}
a_v^{(t+1)}-a_v^{(t)}=-\eta\frac{\partial\ell_C^{(t)}}{\partial a_v^{(t)}}=-\eta\left(p_v^{(t)}-\mathbf{1}[v=C]\right),\qquad \eta>0,
\label{eq:av}
\end{equation}
where the training loss is $\ell_C^{(t)}=-\ln p_C^{(t)}$. $a_C^{(t)}$ is raised by $\eta(1-p_C^{(t)})$ and every other logit is lowered by $\eta p_v^{(t)}$. For any $u\notin\mathcal{S}$, the ratio $z(t)=p_C^{(t)}/p_u^{(t)}$ satisfies
\begin{equation}
z(t+1)=z(t)\exp\left[\eta\left(1-p_C^{(t)}+p_u^{(t)}\right)\right]>z(t).
\label{eq:zt}
\end{equation}
Therefore, under repeated updates on the same context and seen target, the probability ratio of the seen continuation $C$ relative to any unseen continuation $u$ increase. In particular, validation-only continuations $u\in\mathcal{V}$ become progressively less probable relative to $C$. This provides an idealized mechanistic explanation for the crowding-out effect described in Section~\ref{sec:sharpening}.

\subsection{From novel continuations to the frequency law}
\label{app:coverage}

This subsection derives the affine relation of Section~\ref{sec:coverage} (Equation~\eqref{eq:coverage-main}). Fix a training step $t$ and group prediction positions by the training count $F\ge1$ of their context. Within a frequency bin, let $f(F)$ be the fraction of validation positions whose true continuation never followed that context in training. The token itself may be common; what training never showed is the context--continuation pair. The gap splits exactly into two terms:
\begin{equation}
\begin{aligned}
\gap_t(F)&=f(F)D_t(F)+\big(1-f(F)\big)S_t(F),\\
D_t(F)&=\ell^{\mathrm{val}}_{\mathrm{novel},t}(F)-\ell^{\mathrm{tr}}_t(F),\qquad
S_t(F)=\ell^{\mathrm{val}}_{\mathrm{seen},t}(F)-\ell^{\mathrm{tr}}_t(F).
\end{aligned}
\label{eq:gap-decomp}
\end{equation}
Here $D_t$ is the excess loss of novel continuations over the training loss (the novel-continuation penalty of Section~\ref{sec:sharpening}) and $S_t$ is the corresponding seen term; both compare a validation loss with the same online training loss. All three losses are bin means at the same logged step and are not matched context by context. Coverage sets the mixture weights, and the model determines the two loss differences.

If both differences vary little with $F$, the decomposition is approximately affine in the novel fraction:
\begin{equation}
\gap_t(F)\approx A_t f(F)+B_t.
\label{eq:gap_j}
\end{equation}
In the constant-penalty approximation of Equation~\eqref{eq:coverage-main}, $A_t=a_t-b_t$ and $B_t=b_t$. When the seen term is also negligible, it reduces to $\gap_t(F)\approx D_t f(F)$: the gap is the fraction of positions exposed to crowding-out times their loss penalty. The affine relation describes the measured bins well (Figure~\ref{fig:coverage}b,c). Its intercept $B_t$ need not equal the seen term, because frequency-dependent components can also produce a close fit (Appendix~\ref{app:coverage-tests}).

Good--Turing estimation connects the missing continuation mass to singleton counts in the corpus (Appendix~\ref{app:gt-identity}). For a fixed context with a conditional tail $p_{i,(r)}\sim A_i r^{-s}$, the expected missing mass decays as $F^{-(1-1/s)}$ over an intermediate range (Equation~\eqref{eq:occupancy}), which gives $f$ a slow, approximately power-law decay. Pooling heterogeneous contexts and a nonzero intercept both change the slope that the gap inherits (Equation~\eqref{eq:coverage-effective-exponent}).

\begin{figure}[!htbp]
\centering
\includegraphics[width=\linewidth]{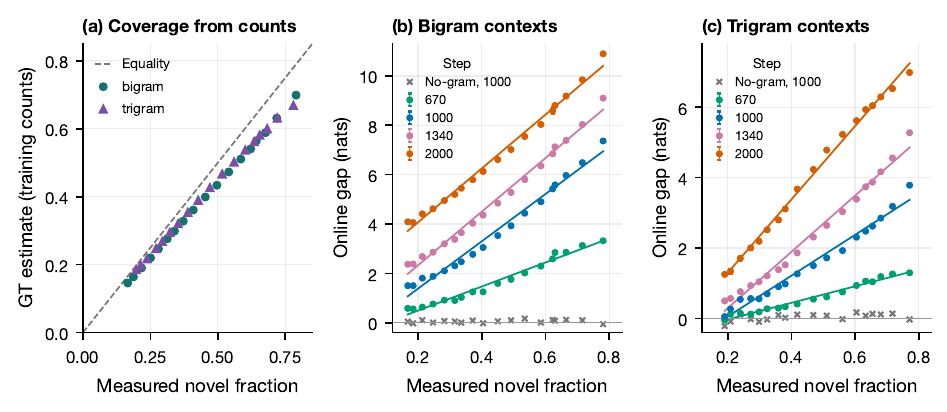}
\caption{Coverage and the gap. (a) Good--Turing estimate from training counts (Equation~\eqref{eq:gt-bin}) versus the measured novel fraction; dashed line: equality. (b,c) Online gap versus measured novel fraction with affine fits at steps 670, 1000, 1340 and 2000; grey crosses show the no-gram control at step 1000. Points are frequency bins; bars show standard deviations across seeds 42, 43 and 44.}
\label{fig:coverage}
\end{figure}

\subsection{Good--Turing estimation and continuation coverage}
\label{app:gt-identity}
Good--Turing estimation relates unseen probability mass to observed singletons \citep{good1953population,gale1995good}. For context $i$, let $p_{iy}$ be its continuation distribution and $m_i(F)=\sum_y p_{iy}(1-p_{iy})^F$ its expected missing mass after $F$ independent draws; this is the expected novel-continuation rate $\mathbb{E}[f_i(F)]$ that Equation~\eqref{eq:occupancy} approximates. If $K_{i,1}(k)$ counts continuation types occurring exactly once in $k$ draws, then
\begin{equation}
\frac{\mathbb E K_{i,1}(F+1)}{F+1}
=\sum_y p_{iy}(1-p_{iy})^F=m_i(F).
\label{eq:gt-single-context}
\end{equation}
This is an expectation identity for a fixed context. The unshifted ratio $K_{i,1}(F)/F$ has expectation $m_i(F-1)$ and equals one at $F=1$ by construction. Pooling the contexts observed $F$ times requires a model of how context counts arise, given next.

\paragraph{Heterogeneous contexts and token weighting.}
Assume training context counts $N_i$ are Poisson with means $\lambda_i=Tq_i$, continuations are conditionally independent, and an independent validation token has context distribution $q_i$ and the same conditional probabilities $p_{iy}$. Write $\pi_F(\lambda)=e^{-\lambda}\lambda^F/F!$. Conditional on the validation context having training count $F$, its population novelty probability is
\begin{equation}
f_\star(F)=\frac{\sum_i q_i\pi_F(\lambda_i)m_i(F)}{\sum_i q_i\pi_F(\lambda_i)}.
\end{equation}
Define $n(F)=\sum_i\mathbf1[N_i=F]$ and $N_1(F)=\sum_i\mathbf1[N_i=F]K_{i,1}(F)$. Using $(F+1)\pi_{F+1}(\lambda)=\lambda\pi_F(\lambda)$ gives
\begin{equation}
\begin{aligned}
\mathbb E N_1(F+1)&=T\sum_i q_i\pi_F(\lambda_i)m_i(F),\\
(F+1)\mathbb E n(F+1)&=T\sum_i q_i\pi_F(\lambda_i),\\
f_\star(F)&=\frac{\mathbb E N_1(F+1)}{(F+1)\mathbb E n(F+1)}.
\end{aligned}
\label{eq:gt-poisson}
\end{equation}
This is a ratio of expectations under the sampling model; the corresponding ratio of observed counts is a plug-in estimate of it. For a bin $B$ of exact counts, the train-only estimator is
\begin{equation}
\widehat f_{\mathrm{GT}}(B)=\frac{\sum_{F\in B}N_1(F+1)}{\sum_{F\in B}(F+1)n(F+1)}.
\label{eq:gt-bin}
\end{equation}
Its pooling weights use training counts only, so the estimate uses no validation data.

\paragraph{Corpus check.}
We count contexts in training shard 1 (49,716,936 tokens) and measure novelty directly on 49,102,848 prediction positions of held-out shard 2. Training counts are taken over the concatenated token stream; validation contexts follow the model's 2049-token chunks. In the 18 frequency bins with $1\le F\le1000$, Equation~\eqref{eq:gt-bin} gives 0.870--0.885 times the measured novel fraction for bigram contexts and 0.857--0.937 for trigram contexts, with no rescaling fitted on validation data (Figure~\ref{fig:coverage}a). At $F=1$ the measured novel fractions are 0.790 (bigram) and 0.780 (trigram), against estimates of 0.699 and 0.669. The estimator therefore tracks the frequency dependence of novelty closely but underestimates its level. A similar underestimate appears when the training shard itself is split in two, so a shift between the training and validation shards is not its only source; natural text also departs from the Poisson and independence assumptions.

\subsection{Position-class decomposition of the excess validation loss}
\label{app:novel-share}

Table~\ref{tab:novel-share} decomposes the both-table model's excess validation loss over the no-gram control (seed 42, step 1000) by whether the true continuation followed its context in the training shard. Contexts are counted at bigram or trigram granularity; \emph{novel} includes positions whose context itself is absent from training ($F=0$). The 589,248 classified positions are the 589,824 fixed validation positions minus the first two predictions in each of the 288 validation sequences. Probabilities in these checkpoint probes are floored at $5.96\times10^{-8}$, so the losses are censored.

\begin{table}[!htbp]
\centering\small
\caption{Share of the step-1000 excess validation loss ($1.675$ nats per position, 589{,}248 positions) carried by each position class. Signed contributions sum to the total within either context granularity.}
\label{tab:novel-share}
\begin{tabular}{llrr}
\toprule
context & class & share of positions & share of excess loss \\
\midrule
bigram & novel continuation & 31.26\% & 72.45\% \\
bigram & seen continuation & 68.74\% & 27.55\% \\
trigram & novel continuation & 65.67\% & 102.02\% \\
trigram & seen continuation & 34.33\% & $-2.02$\% \\
\bottomrule
\end{tabular}
\end{table}

The trigram novel share exceeds 100\% because the seen class contributes slightly negatively: on continuations seen after their trigram context, the both-table model has lower validation loss than the control. These shares refer to the excess validation loss over the no-gram control, which differs from the within-model train--validation gap of Equation~\eqref{eq:gap-decomp}. In particular, they include the $F=0$ positions, which the frequency-bin analysis excludes because no training counterpart exists.

\subsection{Tests of the coverage approximation}
\label{app:coverage-tests}
\label{app:gt-fits}
\paragraph{Protocol.}
The online-gap tests use the both-table model and the no-gram control with seeds 42, 43 and 44 in the standard input-injection setting: RMSProp tables without momentum, a table learning rate $128\times$ the backbone's, $2^{20}$ rows per table, backbone learning rate $0.0006$, and a 100-step warmup followed by a constant learning rate. Training replays shard 1; the fixed validation pool contains 589,824 positions from the first four batches of shard 2. The novel fraction of each bin is counted on the same validation positions that enter its loss. Training losses are recorded before the update and validation losses after it, at the same logged step.

We use all 18 frequency bins with $1\le F\le1000$; each has at least 300 training and validation tokens at the reported steps, and bins with negative gaps are kept. The bins are $1,2,3,4,5,6$--$8,9$--$12,13$--$20,21$--$30,31$--$50,51$--$75,76$--$100,101$--$150,151$--$200,201$--$300,301$--$500,501$--$750,751$--$1000$. Steps 670, 1000, 1340 and 2000 lie near the ends of epochs 2, 3, 4 and 6. We fit the three-seed mean gap by least squares with equal weight per bin; the same held-out pool supplies $f$ for all seeds. Figure~\ref{fig:coverage} uses linear axes throughout and shows the no-gram comparison at step 1000.

\paragraph{Power-law comparison.}
To compare the frequency dependence of gap and novelty, we fit $\log y=\log c-\gamma\log F$ with equal weight on the same 18 bins, representing each bin by the geometric mean of its endpoints. At step 1000 all three-seed mean gaps are positive, so no bins are removed for the log fit. Table~\ref{tab:coverage-exponents} reports both branches. The bigram exponents are close, $0.251$ versus $0.229$, whereas the trigram gap decays more steeply and departs more from a power law. These are finite-range descriptive fits, not universal corpus exponents; the conditional-tail argument in Equation~\eqref{eq:occupancy} does not by itself fix the exponent of a cross-context fit.

\begin{table}[!htbp]
\centering\small
\caption{Matched power-law fits at step 1000. Gap uses the three-seed mean; novelty is measured directly on the shared fixed validation pool. Both use the same frequency bins and equal-bin log--log regression.}
\label{tab:coverage-exponents}
\begin{tabular}{lrrrr}
\toprule
Context & Gap exponent $\beta$ & $R^2_{\log}$ & Novelty exponent $\alpha$ & $R^2_{\log}$\\
\midrule
Bigram & 0.251 & 0.997 & 0.229 & 0.981\\
Trigram & 0.466 & 0.843 & 0.212 & 0.985\\
\bottomrule
\end{tabular}
\end{table}

\paragraph{Fit quality.}
For fits to the three-seed mean at the four reported steps, relative rms residuals are 3--8\% for bigram and 3--11\% for trigram contexts (Table~\ref{tab:affine}). The intercept matters: at step 2000 the bigram fit has $B_t=1.89$ nats, and forcing the line through zero increases the rms error from 0.20 to 0.76 nats. The proportional form $\gap_t\approx D_tf$ therefore fits the later gap much less well than the affine form.

\begin{table}[!htbp]
\centering\small
\caption{Online gap fit $A_tf(F)+B_t$ to the three-seed mean over 18 bins with $1\le F\le1000$. Coefficients and rms errors are in nats; relative rms divides by the mean absolute bin gap. The holdout column fits alternating frequency bins and evaluates on the others.}
\label{tab:affine}
\begin{tabular}{llrrrrr}
\toprule
Context & Step & $A$ & $B$ & rms & Relative & Holdout\\
\midrule
Bigram & 670 & 4.87 & $-0.48$ & 0.137 & 8.2\% & 0.169\\
 & 1000 & 9.58 & $-0.53$ & 0.233 & 6.3\% & 0.269\\
 & 1340 & 10.86 & 0.16 & 0.205 & 4.1\% & 0.228\\
 & 2000 & 10.90 & 1.89 & 0.202 & 3.0\% & 0.233\\
Trigram & 670 & 2.35 & $-0.50$ & 0.063 & 10.7\% & 0.060\\
 & 1000 & 5.81 & $-1.11$ & 0.155 & 9.9\% & 0.132\\
 & 1340 & 8.00 & $-1.31$ & 0.172 & 7.2\% & 0.180\\
 & 2000 & 10.42 & $-0.79$ & 0.129 & 3.2\% & 0.132\\
\bottomrule
\end{tabular}
\end{table}

Weighting bins by validation tokens changes the relative rms to 3.3--10.5\% (bigram) and 3.2--9.1\% (trigram). The residuals are structured and larger than the seed-to-seed variation. A close affine fit also does not require constant $D_t$ and $S_t$: in a fixed-probe diagnostic of the seed-42 both-table model at step 1000, the bigram fit has a negative intercept although every measured seen term is positive, because the frequency dependence of $D_t$ and $S_t$ compensates.

\paragraph{Exact-count comparisons.}
Fixed-probe losses of the seed-42 both-table model at step 1000 allow a comparison at each exact count $F=2,\ldots,8$. We stratify contexts by their training singleton fraction into $[0,.25)$, $[.25,.5)$, $[.5,.75)$ and $[.75,1]$ and keep cells with at least 200 training and validation positions. At bigram $F=4$, the lowest and highest strata have $f=0.247$ and $0.795$ and gaps of 3.234 and 6.013 nats. At trigram $F=4$, the gap rises from 1.448 to 2.619 nats over the first three strata and falls slightly to 2.505 in the highest. At a fixed count, contexts with more singleton continuations therefore tend to have larger gaps. Singleton fraction is not randomized and also reflects conditional entropy. These probe losses use probabilities floored at $5.96\times10^{-8}$ (at most 1,094 zeros per array), so they are censored; the novelty counts are unaffected by the floor.

\paragraph{Corrections to the leading power law.}
Writing $D_t(F)=a_t+\delta D_t(F)$ and $S_t(F)=b_t+\delta S_t(F)$ in the exact decomposition gives
\begin{equation}
\gap_t(F)=(a_t-b_t)f(F)+b_t
+f(F)\delta D_t(F)+[1-f(F)]\delta S_t(F).
\label{eq:coverage-corrections}
\end{equation}
If $f(F)\approx cF^{-\alpha}$, the first term is a power law. The seen-term offset and the frequency-dependent remainder need not share its exponent or be small. These terms allow coverage to track the gap even when their fitted exponents differ.

\paragraph{Effective exponent.}
If $f(F)=cF^{-\alpha}$ exactly and $\gap=Af+B$ with $A$ and $B$ fixed across $F$, the local exponent of a positive gap is
\begin{equation}
\beta_{\mathrm{eff}}(F)=-\frac{\mathrm d\log\gap}{\mathrm d\log F}
=\alpha\,\frac{Af(F)}{Af(F)+B}.
\label{eq:coverage-effective-exponent}
\end{equation}
It equals $\alpha$ only when the intercept is negligible; a positive intercept flattens the gap and a negative one steepens it. Because $A_t$ and $B_t$ change with training step (Table~\ref{tab:affine}) and with table capacity, the observed gap exponent can change while the corpus stays fixed. These tests cover one corpus, optimizer and injection path, and within that setting support coverage as a contributor to the frequency dependence of the gap.

\section{Mechanistic Experiments and Diagnostics}
\label{app:mechanistic}

\subsection{Replay sharpening and crowding-out}
\label{app:sharpening}
\label{app:replay-clue}

The experiments in this section examine whether replay increasingly favors continuations observed in training, as proposed in Section~\ref{sec:sharpening}. The temporal motivation is that the corpus remains fixed while the gap keeps growing. In the 20-epoch both-table experiment with seed 42, the total online gaps at the ends of epochs 1--4 are $-0.062$, $1.199$, $2.730$ and $3.917$ nats. Their increments are comparable early in training and later decline, reaching $0.241$ nats from epoch 19 to 20. This is a clue about the model's response to replay, rather than evidence for a constant increase in novel-continuation loss.

\subsubsection{Experimental setup and primary observations}
\label{app:sharpening-setup}

We use checkpoint probes to examine individual continuation distributions and fixed-validation evaluation to measure their aggregate behavior. The diagnostics examine models with both bigram and trigram tables and, where available, a no-gram control. All runs discussed here use seed 42. A training epoch contains 337 optimizer updates; the five epoch-end checkpoints are at steps 337, 674, 1011, 1348 and 1685. The input and aggregation conventions of each diagnostic are specified below.

\paragraph{Aggregate validation diagnostics.}
Figure~\ref{fig:subspace-ce} uses full causal prefixes and position-weighted cross-entropy, comparing the original both-table checkpoints with a separately evaluated no-gram control at the same five epoch ends. Of 589{,}248 classified validation positions, 405{,}058 have a training-seen bigram context and continuation, and 158{,}790 have a training-seen context with a novel continuation; the remaining 25{,}400 positions have contexts absent from training and are excluded from the two curves. Over the first three epochs, seen-continuation cross-entropy falls from 4.3 to 2.8 nats for the control and stays near 3.4 for the both-table model. For novel continuations, the control improves from 6.9 to 5.6 nats, whereas the both-table model deteriorates from 7.5 to 9.0. This aggregate contrast supports a novel-continuation penalty under replay without implying that every individual context follows the same trajectory.

Predictive entropy and cross-entropy come from a separate evaluation of another pair of seed-42 both-table and no-gram runs in the same setting, at step 1000, near the end of epoch 3. It uses all 589{,}824 fixed validation positions, including the 576 sequence-initial positions excluded from the classification above. At this checkpoint, the both-table model has entropy 2.93 nats and cross-entropy 5.26, compared with 3.59 and 3.58 for the control. The both-table model is thus more confident despite being less accurate, consistent with excessive concentration on a limited set of continuations \citep{carlsson2025hyperfitting}. These full-prefix measurements and the context-only illustration use different inputs and estimands; they are not a matched per-example loss decomposition.

\FloatBarrier
\paragraph{Context-only continuation probabilities.}
Figure~\ref{fig:actual-logits-examples} compares the no-gram and both-table models on three contexts at the ends of epochs 1 and 5. After \emph{petting your}, \emph{dog} occurs once in training and ten times in the held-out recount, whereas \emph{cat} occurs zero and four times, respectively. During replay, the both-table model assigns more probability to \emph{dog} and less to the novel continuation \emph{cat}; the no-gram model becomes more likely to predict both.

\paragraph{Probe inputs and corpus membership.}
The inputs in Figure~\ref{fig:actual-logits-examples} contain BOS followed by exactly three context tokens. Probabilities use the full 8192-token softmax, and the named animals are single tokens under the training tokenizer. Blue marks continuations seen after that exact context in training, orange marks validation-only continuations, and grey marks continuations absent from both counted splits. Counts use training shard 1 and held-out shards 2--4, counting context--target pairs within 2049-token chunks. A blue token may also occur in validation, as its Val count shows. Colors and counts describe the corpus; the two points for a token are the model's predictions for the same input at epochs 1 and 5. \emph{Others} is defined in the caption.

\subsubsection{Additional contexts and heterogeneous trajectories}
\label{app:actual-logits}

\begin{figure}[!p]
\centering
\includegraphics[width=\linewidth]{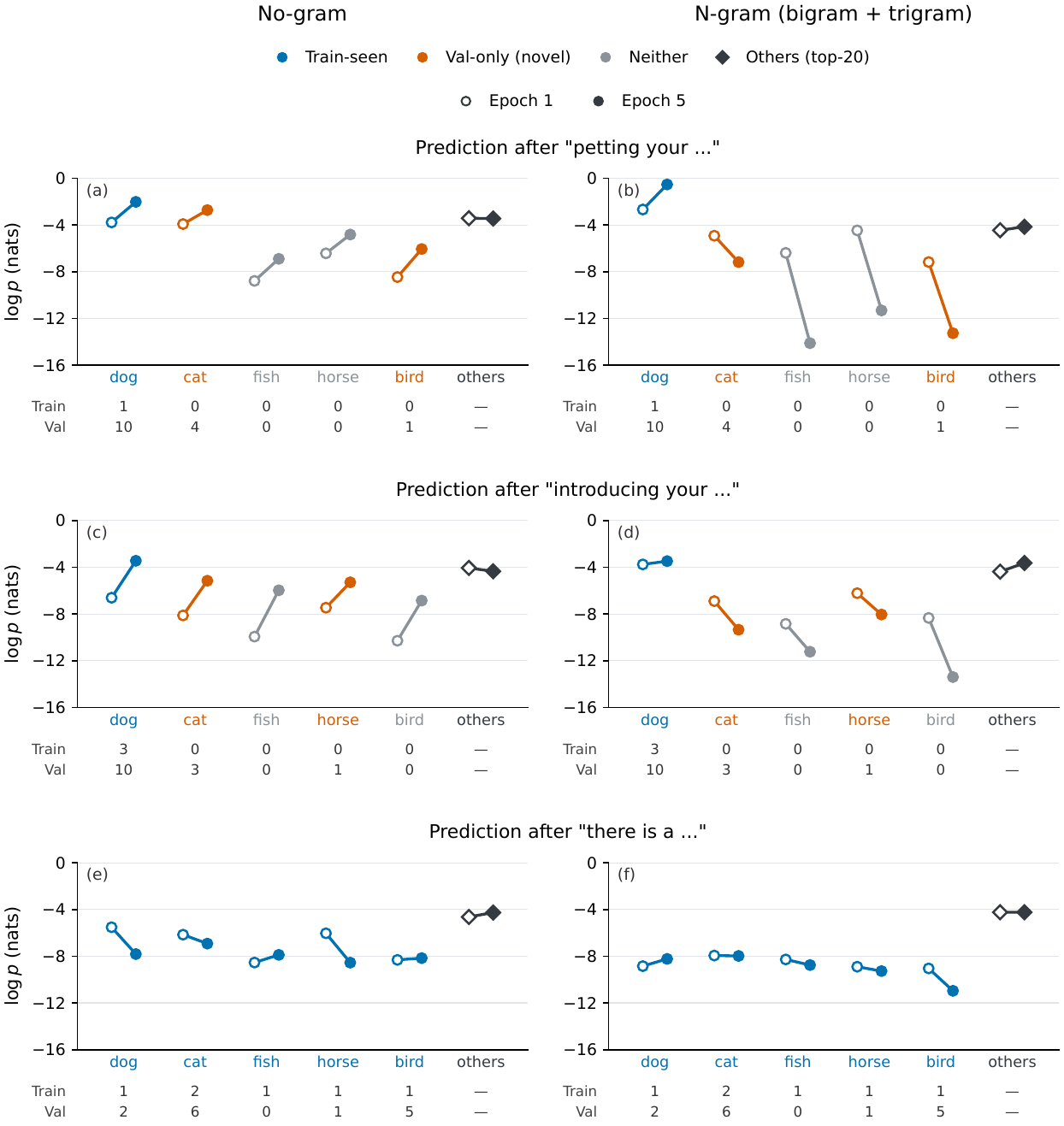}
\caption{Changes in continuation probability from epoch 1 to epoch 5. Rows show \emph{petting your}, \emph{introducing your}, and \emph{there is a}; columns compare no-gram and bigram-plus-trigram memory. Open and filled points mark epochs 1 and 5 (steps 337 and 1685, seed 42); higher $\log p$ means higher probability. Connecting lines join endpoints, without implying monotonic change between them. Blue denotes train-seen continuations, orange validation-only (novel), and grey neither. The Train/Val rows give counts of each exact context--token pair in training shard 1 and held-out shards 2--4; they describe corpus membership, while points show model predictions. Each input is BOS plus three context tokens, using the full 8192-token softmax. Diamonds show the log of the arithmetic mean of the top 20 remaining token probabilities, excluding the named animals and reselected at each checkpoint; this is not total remaining mass.}
\label{fig:actual-logits-examples}
\label{fig:actual-petting}
\label{fig:actual-introducing}
\label{fig:actual-there}
\end{figure}

Figure~\ref{fig:walk-dog} follows one context, \emph{walk their dog}, at each of five epoch ends, showing how its continuation probabilities evolve during replay. Figure~\ref{fig:actual-logits-examples} provides three further examples, comparing the no-gram and both-table models at epochs 1 and 5. The $n$-gram curves in Figure~\ref{fig:subspace-ce} use the same training run and checkpoints as Figure~\ref{fig:walk-dog}, but aggregate full-prefix validation losses across contexts.

The selected \emph{petting your} and \emph{introducing your} contexts come from a cohort where dog follows the context in training and cat does not. Their checkpoint distributions illustrate suppression of an alternative continuation in the $n$-gram model alongside improvement for the no-gram control. The literal \emph{there is a} query has both dog and cat in training; its measured trajectory need not match the schematic in Figure~\ref{fig:replay-transfer}. These examples describe the response of the complete model and do not isolate the effect of training another context.

\begin{figure}[!htbp]
\centering
\includegraphics[width=0.72\linewidth]{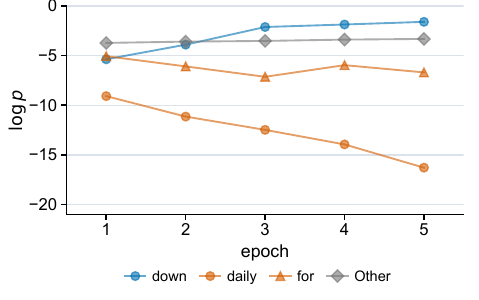}
\caption{Continuation probabilities for \emph{walk their dog} over five replay epochs in the bigram-plus-trigram model (seed 42). Points mark the five epoch-end checkpoints (steps 337, 674, 1011, 1348 and 1685); higher $\log p$ means higher probability. Lines connect successive checkpoints. Blue marks \emph{down}, observed once in training; orange marks \emph{daily} and \emph{for}, each observed once in the counted validation window and absent from training after this context. The input is the three context tokens without BOS, using the full 8192-token softmax. The grey \emph{Other} line shows the log of the arithmetic mean of the top 20 remaining token probabilities, excluding \emph{down}, \emph{daily} and \emph{for} and reselected at each checkpoint; this is not total remaining mass.}
\label{fig:walk-dog}
\end{figure}

\FloatBarrier
\FloatBarrier

\subsubsection{Variation across repeated runs}
\label{app:run-variation}

Our training runs do not enforce deterministic floating-point execution, so fixing the seed does not ensure identical trajectories. In seed-42 reruns, training loss already differed by about $10^{-5}$ nats at step 2. A repeated context-only probe of \emph{walk their dog}, whose training continuation is \emph{down} and whose validation continuations are \emph{daily} and \emph{for}, retained the overall decrease in training-continuation loss and increase in validation-continuation loss, but the intermediate token probabilities were not identical and included reversals. Small numerical differences can therefore accompany different trajectories for an individual continuation, even when the overall trend is preserved.

\FloatBarrier
\subsection{Context locality under loss masking}
\label{app:context-locality}

\subsubsection{Experimental setup and original observations}
\label{app:mask-setup}

The experiment tests whether removing a context's own training loss suppresses further gap growth while the remaining contexts continue training. The original diagnostic selects a random 2\% of training trigram contexts (set $A$) and excludes next-token loss at their positions from the start of epoch 2 or 3. Input tokens and trainable parameters are retained, and the training loss is averaged over the remaining positions. A disjoint 2\% of contexts (set $B$) serves as unmasked peers, and each intervention is paired with normal training from the same checkpoint.

Figure~\ref{fig:subspace-mass} sums full-vocabulary softmax probabilities over training-seen or validation-only continuations, then averages the resulting mass over fixed validation positions. Validation-only continuations follow the context in fixed validation but not in training. The two supports need not exhaust the vocabulary, and their masses use separate vertical scales. Figure~\ref{fig:subspace-gap} shows the cohort's mean fixed-probe gap, $\bar\ell_{\mathrm{val},S}-\bar\ell_{\mathrm{tr},S}$. Weighting each context by its validation-position share $q_i$ turns this into the cohort's contribution to the total gap,
\begin{equation}
C_S=\sum_{i\in S}q_i\bigl[\bar\ell_{\mathrm{val},i}-\bar\ell_{\mathrm{tr},i}\bigr],
\label{eq:context-gap-contribution}
\end{equation}
which is about seventy times smaller because the cohort holds roughly 1.4\% of the validation positions; the reconciliation below uses this weighted form.

The four-epoch diagnostic uses seed 42. Its mass panel compares the both-table and no-gram models with masking from epoch 2, evaluated at steps 337, 674, 1011 and 1348, with 8,167 masked and 8,266 peer validation positions. The gap panel compares normal training with masking from epoch 2 or epoch 3 in the both-table model, plotting the mean fixed-probe gaps of the masked cohort and of its peers. Protocol differences from the separate one-epoch experiments are discussed below.

In the displayed trajectories, the masked cohort's mean gap largely stops growing after the intervention, while the peer and normal-training curves continue to rise. In the $n$-gram model, unmasked contexts gain train-seen mass from $0.43$ to $0.69$, while masked contexts change from $0.43$ to $0.40$. The no-gram control gains train-seen mass with or without masking ($0.22\to0.39$ masked, versus $0.41$ unmasked at epoch 4), and its validation-only mass also increases ($0.013\to0.06$). Removing a context's own replay supervision stops further sharpening in that context, while the other contexts keep sharpening.

\subsubsection{Further analysis and interpretation}
\label{app:mask-reconciliation}

Two separate one-epoch experiments examine whether stable average gap implies stable predictions for individual contexts. They sample 2\% of exact training trigram contexts uniformly without replacement (selection seed 20260919). Of 379{,}789 selected contexts, 5{,}565 have fixed-validation support, covering 8{,}083 positions; a disjoint peer sample has 5{,}571 measurable contexts and 8{,}082 positions. Losses preserve full causal prefixes and are recomputed at a common checkpoint over each context's training and validation occurrences. We report the equally weighted mean of these per-context gaps; the separately computed contribution in Equation~\eqref{eq:context-gap-contribution} instead uses $q_i=N_{\mathrm{val},i}/589{,}824$, without renormalizing within a cohort.

The epoch-2 experiment compares normal and masked training over steps 337--674; the epoch-3 experiment makes the same comparison over steps 674--1011. Both use seed 42. Within each pair, the two conditions begin from the same checkpoint and retain the original batch-token loss denominator. These are separate experiments, not segments of the four-epoch trajectories in Figure~\ref{fig:subspace-evidence}. Their settings follow Appendix~\ref{app:setting}, with two $2^{20}$-row tables, backbone LR $0.0006$ and RMSProp table LR $0.0768$.

For epoch 2, the selected contexts' mean gap increases by $1.018$ nats under normal training and by only $0.007$ with masking; the peer increments remain $0.997$ and $0.998$. For epoch 3, the corresponding selected-context changes are $0.869$ and $-0.065$ nats, while the peer changes are $0.859$ and $0.856$. These comparisons support suppression of average additional gap accumulation. The masked gap can still rise temporarily within an epoch, and its pre-existing value does not disappear.

Average stability does not imply that individual predictions remain fixed. In the masked epoch-3 run, mean training and validation losses both increase, by $0.445$ and $0.380$ nats. Their similar changes produce the small net gap change. The mean absolute per-context gap change is $0.710$ nats in epoch 2 and $0.737$ in epoch 3, much larger than the signed averages; individual contexts move in both directions. Changes also vary across frequency groups, with sparse support at high frequencies. Shared-backbone updates still move individual masked contexts, but they shift training and validation losses together, so the average gap of the masked cohort stays flat.

A variant of the intervention masks 2\% of trigram-table rows instead of context types. In a seed-42 comparison with masking starting from epoch 2, the online masked-cohort gap contribution is $0.0353$ nats at step 670, versus $0.0247$ under normal training. With masking from step 675, the corresponding contributions are $0.0551$ versus $0.0492$ at step 790. These quantities use the online training mean, whereas the figure and paired analyses use fixed probes. Under fixed replay, a single online value depends on where the logged window falls within the epoch, so we also average the online contribution over each epoch, omitting short boundary windows. In the four-epoch runs of Figure~\ref{fig:subspace-evidence}, this epoch mean for the masked cohort is $0.0094$, $0.0076$ and $0.0059$ nats over epochs 2--4 when masking starts at epoch 2, and $0.0277$ and $0.0241$ over epochs 3--4 when it starts at epoch 3; under normal training it rises from $0.0080$ to $0.0263$ to $0.0401$. On this measure, the masked cohort stays close to normal training during its first masked epoch, and the suppression appears in later epochs. The row-masking runs end with their first masked epoch, before this stage, which is consistent with their small differences from normal training.

\section{DeepSeek-Style Replication}
\label{app:deepseek}

\subsection{Experimental goal}
\label{app:deepseek-goal}

Section~\ref{sec:setting} and Appendix~\ref{app:setting} isolate forking in a minimal Transformer with a simple n-gram memory pathway. Here we ask two further questions: does replay-aligned train--validation divergence also occur when memory is embedded in a more structured architecture, and does suppressing low-frequency retrieval still reduce it? We therefore train a compact DeepSeek-style model from scratch, retaining MLA-lite attention, mixture-of-experts (MoE) feed-forward layers, four-stream manifold-constrained hyper-connections (mHC), and Engram retrieval and fusion \citep{cheng2026engram}. This experiment tests the phenomenon in their combined setting.

\subsection{Model architecture and memory pathway}
\label{app:deepseek-architecture}

The model is a 12-layer decoder with hidden width 768. Engram modules act at layers 2 and 6, before the attention and feed-forward sublayers. Each module retrieves bigram and trigram features from the input token sequence and adds their fused representation to the residual streams. In the terminology of this paper, this is multi-layer input over-encoding. Table~\ref{tab:ds-architecture} specifies the architecture used by all three runs.

\begin{table}[!htbp]
\centering\small
\caption{Architecture of the DeepSeek-style replication. Layer indices start at one. These settings are fixed across the full-memory and two retrieval-mask conditions.}
\label{tab:ds-architecture}
\begin{tabular}{@{}p{0.23\linewidth}p{0.72\linewidth}@{}}
\toprule
Component & Configuration\\
\midrule
Decoder and embeddings & 12 layers; hidden width 768; vocabulary 8192; sequence length 2048; separate token and output embeddings; dropout 0.\\
MLA-lite attention & 12 heads; query latent rank 256 and key/value latent rank 128. Per head: 64 non-rotary query/key dimensions, 32 rotary dimensions, and 64 value dimensions; causal attention; RoPE base 10,000.\\
Feed-forward layers & Layer 1: dense SwiGLU with intermediate width 2048. Layers 2--12: 16 routed experts, top-2 selection, and two shared experts; each expert has intermediate width 512.\\
Expert balancing & Sigmoid router scores, normalized over the selected experts; auxiliary-loss-free correction bias with update rate $10^{-3}$.\\
Residual connections & Four mHC streams with learned pre-, post-, and residual mixing; 20 Sinkhorn iterations. RMSNorm is used in latent attention, memory fusion, and the output pathway.\\
Memory locations & Engram at layers 2 and 6; separate parameters at each layer; bigram and trigram retrieval.\\
Lookup heads and capacity & Eight hash heads per order per layer; 24 dimensions per head, concatenated to width 384 per module. The 32 head capacities are successive distinct primes starting at 393,241 and ending at 393,593.\\
Retrieval keys & Fixed token normalization maps 8192 raw IDs to 6737 lookup IDs; causal multiplicative-XOR hashing. Missing preceding positions use a dedicated padding token. Prediction targets retain the raw vocabulary.\\
Memory fusion & A shared value projection and a separate key projection and context-dependent gate for each residual stream; causal depthwise convolution with kernel size 4 and dilation 3.\\
Parameter counts & 574,761,293 total, including 302,156,736 lookup-table parameters; all trainable at initialization.\\
\bottomrule
\end{tabular}
\end{table}

\paragraph{How memory enters the model.}
The concatenated lookup vectors are projected to a shared value and stream-specific keys. Each gate compares the RMS-normalized key and hidden state of its stream, applies a signed square-root transform to their scaled dot product, and then a sigmoid. The gated value and its causal convolution are added to that stream before attention. Thus the retrieved signal depends on both the local input context and the evolving hidden state, unlike the direct addition of a full-width embedding in the minimal model.

\paragraph{Relation to the minimal setting.}
We keep the hidden width, vocabulary, sequence length, batch size, data replay, and optimization protocol aligned with Appendix~\ref{app:setting}. Relative to the minimal model, the architecture increases depth from 8 to 12 layers and introduces MLA-lite attention, SwiGLU/MoE feed-forward layers, rotary positions, mHC residual connections, and untied embeddings. The memory pathway uses compressed multi-head retrieval with gates and convolution. All architecture settings remain fixed across the three conditions; model and memory capacities differ from those of the minimal model.

\subsection{Data, optimization, and intervention protocol}
\label{app:deepseek-protocol}

\paragraph{Data and training budget.}
We reuse the pretokenized shard-1 corpus of the minimal setting in a fixed order. Its 24,264 packed rows each contain 2049 tokens, yielding 2048 next-token predictions per row. An update uses 72 rows, or 147,456 prediction positions, so one pass takes exactly 337 updates and contains 49,692,672 prediction positions. Training lasts 2000 updates with model and data seed 42, reaching partway through the sixth pass. The eligible validation shards are 2--10 and 6542; evaluation always uses the first four batches of shard 2, totaling 589,824 prediction positions. Reusing this fixed pool removes changes in validation sampling from the curves.

\begin{table}[!htbp]
\centering\small
\caption{Optimization and execution settings for the DeepSeek-style runs. Learning rates are inherited from the minimal-setting protocol, whose selection evidence appears in Appendix~\ref{app:parameter-selection}.}
\label{tab:ds-optimization}
\begin{tabular}{@{}p{0.23\linewidth}p{0.72\linewidth}@{}}
\toprule
Component & Configuration\\
\midrule
Non-table parameters & AdamW; learning rate $6\times10^{-4}$; $(\beta_1,\beta_2)=(0.8,0.95)$; $\epsilon=10^{-8}$. Weight decay 0.1 on matrix and convolution weights, excluding token/output embeddings; biases and one-dimensional parameters have no decay.\\
Lookup tables & Bias-corrected, momentum-free RMSProp; learning rate $0.0768$ ($128\times$ the backbone rate); second-moment decay 0.99; $\epsilon=10^{-10}$; no weight decay.\\
Learning-rate schedule & Both rates increase linearly from $0.25\times$ to $1\times$ over updates 1--100, then remain constant; no warmdown or later decay.\\
Numerics and execution & BF16 forward computation; FP32 model parameters, optimizer states, and cross-entropy logits; activation checkpointing; eager execution; one NVIDIA H200 per run, with the official mHC kernels.\\
Evaluation & Current-batch training loss and fixed validation loss every 10 updates; 200 recorded points per run; no smoothing.\\
\bottomrule
\end{tabular}
\end{table}

The shared optimization settings test whether the established replay regime also produces forking in the new model. The first pass provides a period without replay; continuing for several passes exposes both the onset and persistence of forking. Holding the learning rates constant after warmup also separates replay boundaries from learning-rate changes.

\paragraph{Three comparison conditions.}
The full-memory run retains every retrieval throughout training. Two further runs apply the low-frequency read-mask intervention of Section~\ref{sec:interventions}, here using thresholds $\tau=8$ and $\tau=100$. Both masks remain inactive for the first 337 updates and activate before update 338, the first update of the second pass. For each order $n\in\{2,3\}$, all eight retrieved head vectors are set to zero when the raw input context has training frequency $F_n\leq\tau$, including unseen contexts with $F_n=0$. Frequencies count model-input contexts in one unrepeated training pass, using the same raw-token index and sequence-boundary convention as the minimal-setting intervention. The same rule acts at both Engram layers during training and evaluation, after table lookup and before the key/value projections. All next-token loss terms, table parameters, optimizer states, projection biases, gates, and convolution operations are retained.

\paragraph{Measurement and comparison.}
Each condition is trained independently from initialization with seed 42. Every 10 updates, we record the current-batch training loss before the update and the fixed-pool validation loss after it. The online gap is $G(t)=L_{\mathrm{val}}(t)-L_{\mathrm{train}}(t)$, measured in nats.

\subsection{Observed forking and response to masking}
\label{app:deepseek-results}

\begin{figure}[!htbp]
\centering
\includegraphics[width=\linewidth]{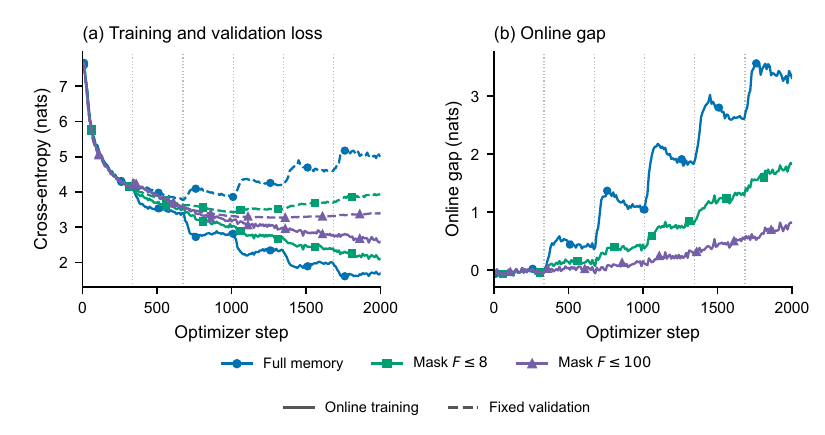}
\caption{Forking and low-frequency retrieval masking in the DeepSeek-style model. Left: current-batch training loss before each logged update (solid) and fixed validation loss after it (dashed). Right: their difference. Dotted lines mark completed 337-update passes; masks activate before update 338. All three independent seed-42 runs complete 2000 updates; curves join the values logged every 10 updates without smoothing. Low-frequency masking attenuates the replay-associated gap growth and lowers final validation loss.}
\label{fig:deepseek}
\end{figure}

\paragraph{Forking recurs under replay.}
Figure~\ref{fig:deepseek} shows that the full-memory model initially improves both training and validation loss. Once the data repeat, training loss separates from validation loss. Validation loss continues to improve during the second pass, reaching its lowest recorded value near update 670, before rising in later passes. These later passes produce the staircase pattern seen in the minimal setting: training loss drops near replay boundaries while validation loss rises, causing repeated increases in the gap. At update 2000, training loss is 1.689 nats and validation loss is 5.005 nats, giving a gap of 3.316 nats. Forking therefore remains observable with multi-layer, gated Engram retrieval in this combined architecture.

\paragraph{Low-frequency masking attenuates the divergence.}
At update 2000, masking $F\leq8$ reduces the gap to 1.835 nats, and masking $F\leq100$ reduces it to 0.817 nats. These are reductions of 44.7\% and 75.3\% relative to full memory. The smaller gaps accompany lower absolute validation losses, by 1.067 and 1.600 nats respectively, as well as higher training losses. Thus the intervention both weakens fitting to replayed training data and improves held-out prediction in these runs. It also allows validation loss to continue decreasing over the first few replay epochs.

\FloatBarrier
\subsection{Conclusions and evidence limits}
\label{app:deepseek-conclusions}

The experiment supports two conclusions within this small-data replay regime. First, forking extends beyond the minimal additive-memory Transformer to the tested DeepSeek-style architecture. Second, its response to low-frequency read masking is consistent with the minimal-setting evidence that access to rare contexts contributes to replay-associated overfitting. Forking persists with the richer retrieval and fusion pathway in this configuration.

We also attempted a no-gram control using the same backbone architecture and optimization settings. The training loss became non-finite during the first pass, and this failure recurred in repeat runs. We therefore could not obtain a completed no-gram baseline of our model for comparison.

\section{Continued-Training and Replay Stress-Test Protocol}
\label{app:sft}

The trainable- and frozen-table runs in Section~\ref{sec:transfer} start from the checkpoint-materialization rerun of the full-pool $20\times$ trigram experiment, with endpoint gap $0.214$ at step 20,220. The original scientific run has the same setting and endpoint gap $0.220$ but does not retain final weights. We restart the backbone AdamW and table RMSProp states and apply a fresh 100-step warmup.

The model is the vanilla 8-layer, 6-head, 768-dimensional NanoGPT with input trigram injection and a clean $2^{20}$-row table. We use BF16 forward computation, no compilation, backbone learning rate $6\times10^{-4}$, table learning rate $0.0768$, and fixed validation. Each continued-pretraining source has 333 batches and is run for five epochs (1665 updates); logs are recorded every 10 steps and at the final boundary. Each batch holds 72 sequences with 2048 prediction targets, and validation uses four fixed batches. The online gap is $G=L_{\mathrm{va}}-L_{\mathrm{tr}}$, where training loss is measured on the current batch before the update and validation loss is measured after the update on fixed batches. In the frozen arm the table receives no updates from the first continued-training step, while the backbone remains trainable.

The ID source continues the pretraining corpus: it trains on shard 21, which the base model never trained on, and validates on shards 22--23. The code source is \href{https://huggingface.co/datasets/ise-uiuc/Magicoder-OSS-Instruct-75K}{Magicoder-OSS-Instruct-75K}~\citep{wei2024magicoder} and the math source is OpenR1-Math-220k~\citep{openr12025math}. Their records are tokenized in source order and packed into one stream; the first 333 batches form the training pool and the next four batches the validation pool. Neither pool is cycled, and one source record in each dataset straddles the boundary between the two pools. All sources are encoded with the tokenizer used in pretraining.

ID uses downstream seed 42. Magicoder, OpenR1, and RLVR-MATH (below) each use downstream seeds 42, 43, and 44 from the same pretrained checkpoint, and the tables report their means. The three seeds share the checkpoint and the data order, so their spread measures run-to-run variation in continued training only; the standard deviation of the endpoint gap is at most 0.015 nats for every source and arm. Figure~\ref{fig:sft} shows the full trajectories, with one-standard-deviation bands where three seeds exist. The endpoint measurements are:

\begin{table}[!htbp]
\centering\small
\caption{Endpoint measurements for continued language-model pretraining from the low-gap checkpoint. All values are nats at step 1665. ID uses seed 42; Magicoder and OpenR1 average downstream seeds 42--44. The table-trainable and table-frozen runs differ only in whether the trigram lookup table receives gradients.}
\label{tab:sft-results}
\begin{tabular}{llrrr}
\toprule
Source & Table & Train & Validation & Gap\\
\midrule
ID & Trainable & 1.913 & 4.376 & 2.463\\
ID & Frozen & 3.030 & 3.372 & 0.341\\
Magicoder & Trainable & 0.619 & 1.543 & 0.924\\
Magicoder & Frozen & 1.103 & 1.301 & 0.198\\
OpenR1 & Trainable & 0.688 & 1.366 & 0.678\\
OpenR1 & Frozen & 1.039 & 1.197 & 0.158\\
\bottomrule
\end{tabular}
\end{table}

These runs optimize the language-model loss on all tokens. They are continued-pretraining measurements, not response-only instruction-SFT results.

\paragraph{Static RLVR-like proxy.}
We additionally ran continued-pretraining runs on \href{https://huggingface.co/datasets/allenai/RLVR-MATH}{RLVR-MATH} from the Tulu 3 release~\citep{lambert2024tulu3}, which contains the 7,500 MATH training problems formatted for RLVR. The prompts carry fixed 4-shot exemplar prefixes that are byte-identical across examples, and the reference-answer field is the reference final answer. Messages were retained verbatim and the reference answer was appended once after an ``Answer:'' marker. The first 95\% and last 5\% of source examples (in original row order) were used as disjoint train and validation pools, then cycled to 333 train batches and 4 validation batches. Because of this cycling, one nominal epoch covers the raw training token stream about 8.54 times, or 42.72 times over five epochs.

The shared setting matches the runs above: downstream seeds 42--44, 1665 updates, five replay epochs, fresh optimizer state and 100-step warmup, input trigram-only clean $2^{20}$-row table, and fixed validation. In the frozen arm only the table is fixed; the backbone remains trainable. Endpoint measurements are:
\begin{table}[!htbp]
\centering\small
\caption{Static RLVR-like verifier-selected continued-pretraining endpoints. All values are nats at step 1665, averaged over downstream seeds 42--44.}
\label{tab:rlvr-results}
\begin{tabular}{llrrr}
\toprule
Source & Table & Train & Validation & Gap\\
\midrule
RLVR-MATH & Trainable & 0.009 & 0.518 & 0.509\\
RLVR-MATH & Frozen & 0.014 & 0.447 & 0.434\\
\bottomrule
\end{tabular}
\end{table}

This is a static verifier-selected proxy, not on-policy reinforcement learning.

\end{document}